\documentclass[11pt]{article}

\usepackage[preprint]{acl}

\usepackage{times}
\usepackage{amsmath}
\usepackage{booktabs}
\usepackage{amssymb}
\usepackage{pifont}
\usepackage{stfloats}
\usepackage{multicol}
\usepackage{float}
\usepackage{graphicx}
\usepackage{calc}   
\usepackage{subcaption}

\usepackage{multirow}
\usepackage{array}
\newcommand{\PreserveBackslash}[1]{\let\temp=\\#1\let\\=\temp}

\newcolumntype{C}[1]{>{\PreserveBackslash\centering}p{#1}}
\newcolumntype{R}[1]{>{\PreserveBackslash\raggedleft}p{#1}}
\newcolumntype{L}[1]{>{\PreserveBackslash\raggedright}p{#1}}
\usepackage{diagbox}
\usepackage[noend]{algpseudocode}
\usepackage[ruled,linesnumbered,lined,commentsnumbered]{algorithm2e}
\usepackage{booktabs}
\usepackage{amsfonts}
\usepackage{bbm}
\usepackage{mathtools}
\usepackage{enumitem}

\usepackage[T1]{fontenc}

\usepackage[utf8]{inputenc}

\usepackage{microtype}

\usepackage{inconsolata}

\title{\textit{Language on Demand, Knowledge at Core}: Composing LLMs with Encoder-Decoder Translation Models for Extensible Multilinguality}

\author{
    Mengyu Bu\textsuperscript{\rm 1,2,3},
    Yang Feng\textsuperscript{\rm 1,2,3}\footnotemark[2] \\
    \textsuperscript{\rm 1}{Key Laboratory of Intelligent Information Processing, Institute of Computing Technology,} \\ Chinese Academy of Sciences (ICT/CAS) \textsuperscript{\rm 2} {State Key Laboratory of AI Safety,} \\ Institute of Computing Technology, Chinese Academy of Sciences \\
    \textsuperscript{\rm 3} {University of Chinese Academy of Sciences, Beijing, China} \\
    \texttt{\href{mailto:bumengyu23z@ict.ac.cn}{bumengyu23z@ict.ac.cn},  \href{mailto:fengyang@ict.ac.cn}{fengyang@ict.ac.cn}}
}

\begin{document}
\maketitle

\renewcommand{\thefootnote}{\fnsymbol{footnote}} 
\footnotetext[2]{Corresponding author: Yang Feng.} 
\renewcommand{\thefootnote}{\arabic{footnote}}

\begin{abstract}


Large language models (LLMs) exhibit strong general intelligence, yet their multilingual performance remains highly imbalanced.
Although LLMs encode substantial cross-lingual knowledge in a unified semantic space, they often struggle to reliably interface this knowledge with low-resource or unseen languages.
Fortunately, pretrained encoder-decoder translation models already possess balanced multilingual capability, suggesting a natural complement to LLMs.
In this work, we propose XBridge, a compositional encoder-LLM-decoder architecture that offloads multilingual understanding and generation to external pretrained translation models, while preserving the LLM as an English-centric core for general knowledge processing.
To address the resulting representation misalignment across models, we introduce lightweight cross-model mapping layers and an optimal transport-based alignment objective, enabling fine-grained semantic consistency for multilingual generation.
Experiments on four LLMs across multilingual understanding, reasoning, summarization, and generation indicate that XBridge outperforms strong baselines, especially on low-resource and previously unseen languages, without retraining the LLM.\footnote{\url{https://github.com/ictnlp/XBridge}}

\end{abstract}
\section{Introduction}

\begin{figure}[t]
  \includegraphics[width=\columnwidth]{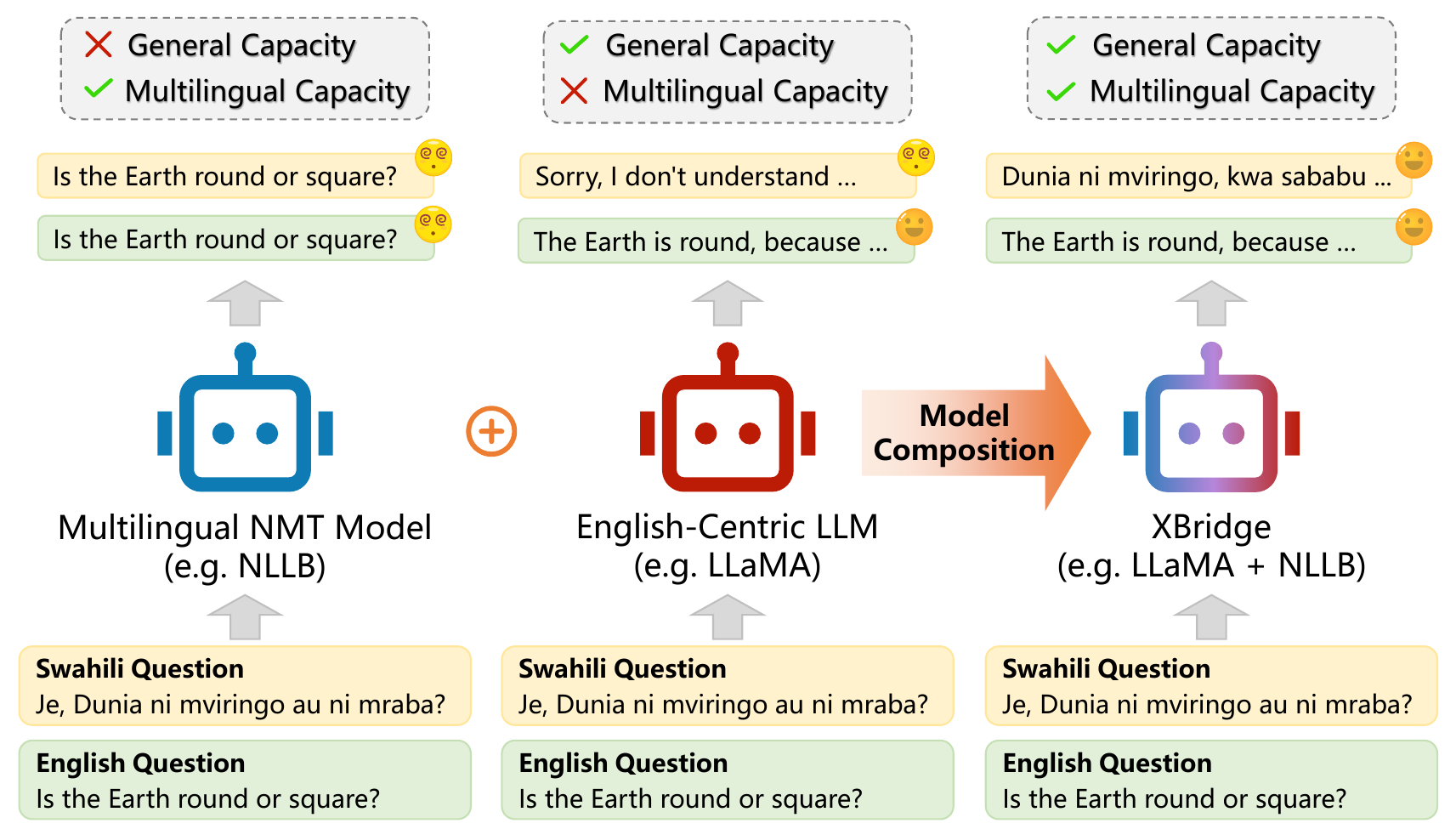}
  \caption{Overview of XBridge. Pretrained multilingual NMT models provide broad language coverage but limited general reasoning capability, while English-centric LLMs excel at general reasoning yet struggle with low-resource or unseen languages. XBridge harmonizes these strengths through model composition, offloading multilingual processing to the pretrained multilingual model while leveraging the LLM as a knowledge core.}
  \label{overview}
\end{figure}

Large language models (LLMs) have demonstrated remarkable general intelligence and reasoning abilities~\cite{touvron2023llama, ustun2024aya, qwen2025qwen25technicalreport}, which are largely grounded in a unified semantic knowledge space.
However, despite possessing substantial cross-lingual knowledge, LLMs exhibit imbalanced multilingual performance: while performing reliably in English and a few high-resource languages, they often fail to robustly understand or generate text in low-resource or unseen languages~\cite{zhu2023extrapolating, chang-etal-2024-multilinguality}.
This suggests that the core limitation of LLMs lies not in the absence of knowledge, but in the difficulty of interfacing this knowledge with diverse linguistic representation spaces.

Fortunately, a wealth of encoder-decoder based neural machine translation (NMT) models~\cite{xue-etal-2021-mt5, nllbteam2022languageleftbehindscaling} specialize in multilingual understanding and generation, and thus provide complementary capabilities to LLMs.
These models support semantic transfer across hundreds of languages, including many low-resource ones, by learning a shared semantic representation space across languages.
In such models, the encoder maps input text from different languages into the shared semantic space, while the decoder subsequently projects these shared representations into target-language outputs.
This closed semantic loop between understanding and generation, along with the modular design of encoder and decoder, naturally complements LLMs.
Realizing such a composition would provide LLMs with extensible multilingual capability, particularly for low-resource or unseen languages that are well modeled by NMT systems but remain challenging for LLMs.

However, existing approaches only partially address this goal, which integrate multilingual encoders to improve multilingual understanding by injecting encoder representations into LLM inputs~\cite{yoon-etal-2024-langbridge, huang2024mindmerger, ruan-etal-2025-layalign}.
While effective for input understanding, these approaches leave generation largely English-centric.
A natural extension is to further incorporate the multilingual decoder, but doing so introduces a fundamental structural challenge.
In NMT, the encoder and decoder are jointly trained within a unified representation space, whereas inserting a frozen LLM in between introduces a transformation from the LLM input space to a different output space shaped by its internal knowledge processing.
Consequently, the LLM outputs no longer match the decoder’s expected cross-attention representations, resulting in semantic misalignment that cannot be resolved by simple projection.

To address this challenge, we propose XBridge, which composes LLMs with pretrained multilingual NMT models for extensible multilinguality.
XBridge adopts an encoder-LLM-decoder architecture, where a multilingual encoder provides robust semantic representations for multilingual inputs, a frozen LLM serves as an English-centric core for knowledge processing, and a multilingual decoder generates outputs in the target language.
From a representation perspective, XBridge constructs a semantic bridge that transforms representations from the multilingual semantic space to the LLM input space, through the LLM output space after knowledge transformation, and finally into the decoder’s generation space.
By explicitly aligning heterogeneous representation spaces across these modules, XBridge resolves the semantic mismatch introduced by inserting a frozen LLM, achieving extensible and generalizable multilingual understanding and generation.

We evaluate XBridge on four LLMs across multilingual understanding, reasoning, summarization, and generation tasks.
XBridge outperforms strong baselines, with significant gains on low-resource and unseen languages while preserving LLM's core capability.
With minimal additional parameters, limited training data, and parameter-efficient training, XBridge brings low-resource and unseen language performance close to that of external NMT models, substantially narrowing the gap across languages without retraining the LLM.
\section{Related Work}

\subsection{Data-Level Multilingual Enhancement for LLMs}

A line of work augments the multilingual capabilities of LLMs at the data level by constructing multilingual training corpora using pretrained multilingual or machine translation models~\cite{li2023bactrian, zhang2023bayling, zhang2024bayling, zhang-etal-2024-enhancing-multilingual}.
Typical approaches translate English instruction into multiple languages~\cite{chen2024breaking}, pre-translate non-English inputs into English before task execution~\cite{qin-etal-2023-cross, chai2025xcot}, or leverage Mix-of-Experts (MoE) for language expansion~\cite{zhang-etal-2025-less}.
Such approaches generally require continual multilingual training of LLMs, which may introduce translation noise and interfere with existing language capabilities.
In practice, balancing performance across high- and low-resource languages remains challenging, as gains on low-resource languages often come at degradation on high-resource ones~\cite{gao-etal-2024-multilingual}.
In contrast, XBridge achieves multilingual generalization through model composition without multilingual retraining of the LLM.

\subsection{Encoder-Augmented Multilingual LLMs}

Another line of work augments LLMs with pretrained multilingual encoders, injecting encoder representations into the LLM to improve multilingual understanding.
\citet{yoon-etal-2024-langbridge} leverage multilingual encoders to support cross-lingual understanding, while \citet{huang2024mindmerger} reintroduce multilingual inputs to better exploit the complementary strengths of language understanding and reasoning in LLMs. \citet{ruan-etal-2025-layalign} further explore layer-wise fusion strategies to enhance the utilization of encoder semantics.
These approaches primarily focus on improving multilingual understanding at the input side, while generation remains governed by the LLM’s native language distribution, typically English.
Moreover, due to differences in training objectives and tokenization schemes, representation gaps persist between multilingual encoders and LLMs, which limit the effective exploitation of encoder semantics.
XBridge differs from prior encoder-augmented methods by additionally incorporating a multilingual decoder to support multilingual generation and by explicitly aligning representations across models, enabling more effective end-to-end multilingual behavior.

\section{Method}

\begin{figure*}[htbp]
  \includegraphics[width=\linewidth]{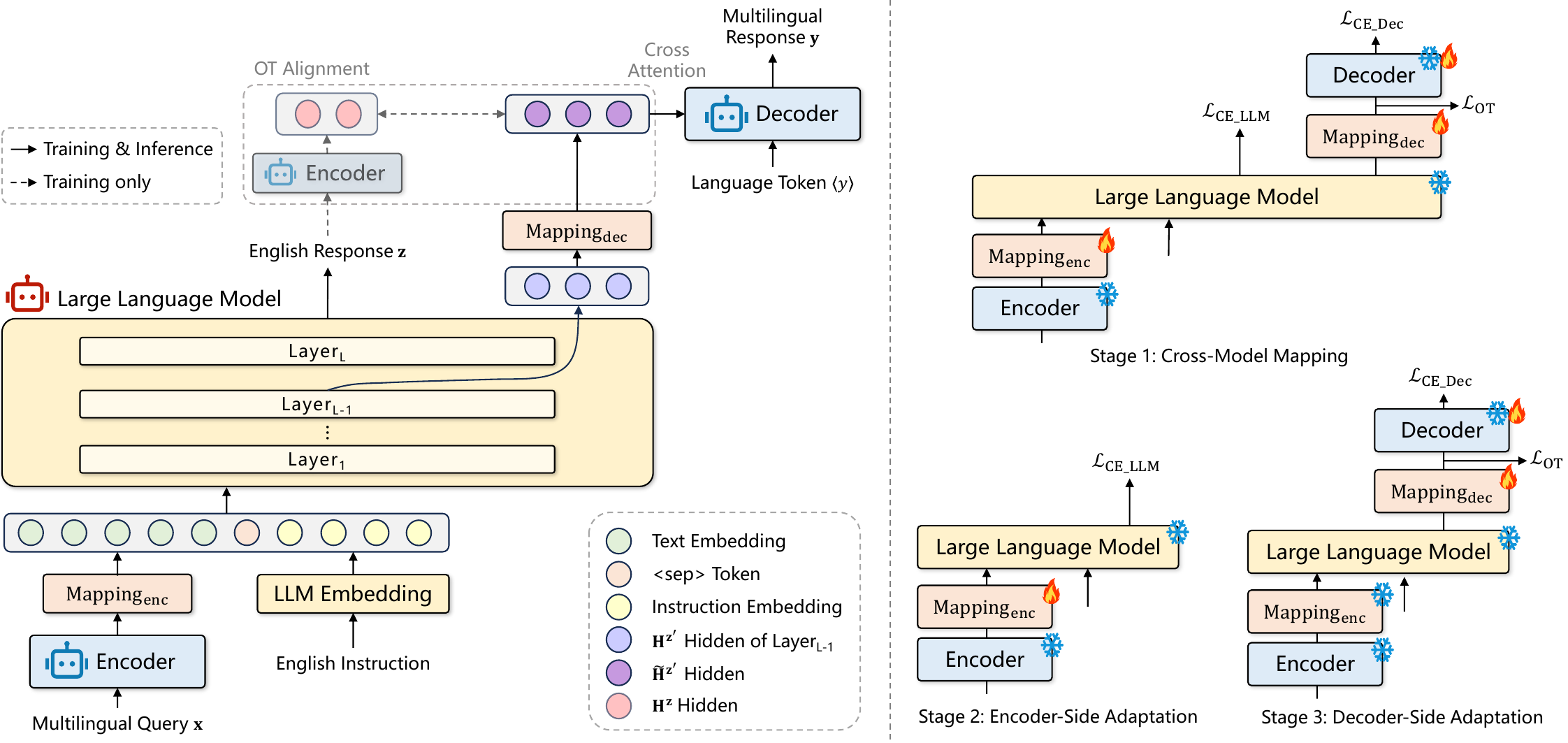}
  \caption{\textbf{Left:} XBridge composes a pretrained multilingual encoder-decoder with an LLM via lightweight mapping layers for multilingual understanding and generation, keeping the LLM frozen as a knowledge core. \textbf{Right:} A three-stage training strategy progressively aligns heterogeneous representations and adapts the encoder and decoder.}
  \label{arch}
\end{figure*}

Figure~\ref{arch} presents the framework of our XBridge, a compositional multilingual framework that integrates a pretrained encoder-decoder NMT model with an LLM. XBridge efficiently offloads multilingual burden to the external NMT model while preserving the LLM as an English-centric core for general knowledge processing.
XBridge adopts an encoder-LLM-decoder architecture, connected by lightweight cross-model mapping layers (Section~\ref{subsection_arch}).
To facilitate fine-grained semantic transfer for multilingual generation, we introduce an optimal transport-based token alignment objective at the LLM-decoder interface (Section~\ref{subsection_ot}).
For stable optimization, XBridge employs a three-stage training strategy that decouples coarse-grained cross-model alignment from task-specific adaptation (Section~\ref{subsection_training}).

\subsection{Architecture}
\label{subsection_arch}

XBridge adopts an encoder-LLM-decoder architecture to compose a pretrained encoder-decoder NMT model with an LLM for extensible multilingual understanding and generation.

Formally, given an input sequence $\mathbf{x} = (x_1,\ldots,x_n)$ in language $L_x$, we first encode it with the pretrained multilingual encoder $\mathtt{Enc}(\cdot)$, producing contextual representations $\mathbf{H}^\mathbf{x} \in \mathbb{R}^{n \times d_e}$.
To bridge the representation gap between the multilingual encoder and LLM, we apply a lightweight mapping $\mathtt{Mapping_{enc}}(\cdot)$ that projects $\mathbf{H}^\mathbf{x}$ into the LLM representation space, yielding $\tilde{\mathbf{H}}^\mathbf{x} \in \mathbb{R}^{n \times d_l}$.
The mapped encoder representations are then injected into the LLM together with a high-resource (English) instruction prompt, enabling the LLM to perform general knowledge processing conditioned on encoder semantics.
Let $\mathbf{z} = (z_1,\ldots,z_m)$ denote the sequence of English tokens generated by the LLM.
Rather than using the final-layer hidden states, we extract the penultimate-layer hidden states, denoted as $\mathbf{H}^\mathbf{z'} \in \mathbb{R}^{m \times d_l}$, as \citet{zhang2025doesalignmentenhancellms} show that the last layer is often tightly aligned with the output vocabulary space, while non-final layers retain richer semantic information.

To support multilingual generation, XBridge further integrates a pretrained multilingual decoder $\mathtt{Dec}(\cdot)$ at the output side.
Specifically, we apply a decoder-side mapping $\mathtt{Mapping_{dec}}(\cdot)$ to project the LLM hidden states into the decoder representation space, obtaining $\tilde{\mathbf{H}}^\mathbf{z'} \in \mathbb{R}^{m \times d_d}$, which are used as key-value representations for cross-attention in the decoder. 
Given target-language tokens $\langle y \rangle$ in language $L_y$ as decoder inputs, the decoder generates the output sequence $\mathbf{y}$ by attending to $\tilde{\mathbf{H}}^\mathbf{z'}$, producing text that follows the target-language distribution while remaining semantically grounded in the LLM’s knowledge processing results.

\subsection{Optimal Transport-Based Alignment}
\label{subsection_ot}

Although the mapped LLM representations $\tilde{\mathbf{H}}^\mathbf{z'}$ can be directly used as cross-attention inputs for multilingual decoding, token-level semantic misalignment may arise due to heterogeneous tokenizations and representation spaces across models.
To encourage fine-grained semantic consistency at the LLM-decoder interface, we introduce an optimal transport (OT)-based alignment objective.

Specifically, given the English token sequence $\mathbf{z} = (z_1,\ldots,z_m)$ generated by the LLM, we re-encode it using the same multilingual encoder $\mathtt{Enc}(\cdot)$, obtaining encoder representations $\mathbf{H}^\mathbf{z} \in \mathbb{R}^{k \times d_e}$, where the sequence length $k$ may differ from $m$ due to heterogeneous tokenizers.
Since $\mathbf{H}^\mathbf{z}$ and the decoder-side LLM representations $\tilde{\mathbf{H}}^\mathbf{z'}$ are both derived from the same LLM output, they are semantically equivalent in expectation, despite residing in different representation spaces.
We therefore align $\mathbf{H}^\mathbf{z}$ with $\tilde{\mathbf{H}}^\mathbf{z'}$ to enforce token-level semantic alignment.

Due to sequence length mismatch by heterogeneous tokenizers, we formulate the alignment as an optimal transport problem~\cite{peyre2019computational}, which computes a soft, many-to-many matching between the two sequences.
Concretely, we define the OT distance between $\mathbf{H}^\mathbf{z}$ and $\tilde{\mathbf{H}}^\mathbf{z'}$ as:
\begin{equation}
\label{eq:ot}
\begin{aligned}
\mathcal{D}^*(\mathbf{H}^\mathbf{z}, \tilde{\mathbf{H}}^\mathbf{z'})
= \min_{\mathbf{T} \geq 0} \sum_{i,j} \mathbf{T}_{ij} \, c(H^{\mathbf{z}}_i, \tilde{H}^{\mathbf{z}'}_j), \\
\text{s.t. } \sum_{j=1}^{m} \mathbf{T}_{ij} = m_i^\mathbf{z}, \quad \forall i \in \{1,\ldots,k\}.
\end{aligned}
\end{equation}
where $\mathbf{T}_{ij}$ denotes the transport mass from $H^{\mathbf{z}}_i$ to $\tilde{H}^{\mathbf{z}'}_j$, and $c(\cdot,\cdot)$ is the transport cost computed using cosine distance.
The mass distribution $\{m_i^\mathbf{z}\}$ is obtained by normalizing $\mathbf{H}^\mathbf{z}$.
Appendix \ref{ot_algorithm} presents details of the OT formulation and optimization.

The OT loss provides flexible, token-level supervision that is robust to length mismatch.
By regularizing the decoder-side mapping with encoder-derived representations of the LLM’s own outputs, the OT objective encourages $\tilde{\mathbf{H}}^\mathbf{z'}$ to preserve semantic structures compatible with the multilingual encoder-decoder space.
This alignment not only improves multilingual generation quality, but also indirectly facilitates more effective utilization of multilingual encoder signals by the LLM.

\subsection{Three-Stage Training Strategy}
\label{subsection_training}

To ensure stable optimization across models and objectives, XBridge employs a three-stage training strategy that progressively aligns heterogeneous representations and adapts the model to downstream tasks, keeping the LLM frozen throughout.

\paragraph{Stage 1: Cross-Model Mapping}

Due to the substantial representation gaps between the multilingual encoder and the LLM, as well as between the LLM and the multilingual decoder, directly bridging heterogeneous components is non-trivial.
We therefore first establish coarse-grained semantic alignment among the multilingual encoder, the LLM, and the multilingual decoder using trilingual translation data $(\mathbf{x}, \mathbf{z}, \mathbf{y})$, where $\mathbf{z}$ is an English sequence generated by the LLM.
In this stage, only the encoder-side mapping $\mathtt{Mapping_{enc}}$, the decoder-side mapping $\mathtt{Mapping_{dec}}$, and the decoder cross-attention layers are trained, optimizing the LLM English generation loss, the multilingual decoder generation loss, and the optimal transport alignment loss.
This stage enables the LLM to interpret multilingual encoder representations and allows the decoder to attend to LLM hidden states for multilingual generation.

\paragraph{Stage 2: Encoder-Side Adaptation}

After cross-model semantic alignment is established, the second stage adapts multilingual input representations to downstream instruction-following tasks.
We fine-tune only the encoder-side mapping layer $\mathtt{Mapping_{enc}}$ on task-specific instruction data by optimizing the LLM English generation loss, while keeping all decoder-related components frozen.
This stage teaches the LLM how to use multilingual representations to perform tasks, building upon the aligned representation space learned in stage~1.

\paragraph{Stage 3: Decoder-Side Adaptation}

The third stage focuses on improving multilingual generation quality by adapting the LLM-decoder interface.
We update only $\mathtt{Mapping_{dec}}$ and the decoder cross-attention layers, optimizing the multilingual decoder generation loss together with the optimal transport alignment loss.
Separating this stage from stage~2 avoids conflicts between LLM and decoder objectives: stage~2 first stabilizes the conditional distribution of the LLM outputs, which stage~3 then exploits to enhance decoder performance without degrading task understanding.


\paragraph{Training Objectives}
Given encoder input sequence $\mathbf{x}$ with encoder representations $\mathbf{H}^{\mathbf{x}}$, the LLM-generated English sequence $\mathbf{z}$ with penultimate-layer hidden states $\mathbf{H}^{\mathbf{z'}}$, decoder-mapped representations $\tilde{\mathbf{H}}^{\mathbf{z'}}$, and multilingual decoder output sequence $\mathbf{y}$, the cross-entropy losses of LLM and decoder are defined as:
\begin{equation}
\mathcal{L}_{\text{CE\_LLM}}
= - \log p_{\text{LLM}}(\mathbf{z} \mid \mathbf{x}, \text{inst}).
\end{equation}
\begin{equation}
\mathcal{L}_{\text{CE\_Dec}}
= - \log p_{\text{Dec}}(\mathbf{y} \mid \tilde{\mathbf{H}}^{\mathbf{z'}}, \langle y \rangle).
\end{equation}

Across stages, the overall training objective is:
\begin{equation}
\mathcal{L}
= \lambda_1 \mathcal{L}_{\text{CE\_LLM}}
+ \lambda_2 \mathcal{L}_{\text{CE\_Dec}}
+ \lambda_3 \mathcal{L}_{\text{OT}}.
\end{equation}
where different loss terms are activated depending on the training stage, as illustrated in Figure~\ref{arch}.
\section{Experiment}
\label{main_experiment_text}

\begin{table*}[t]
\centering
\small
\begin{tabular}{l|cccc|cccc|cc}\toprule
\multirow{2}{*}{\textbf{System}} & \multicolumn{4}{c|}{\textbf{Low-Resource Languages}}               & \multicolumn{4}{c|}{\textbf{High-Resource Languages}}              & \multicolumn{2}{c}{\textbf{Average}} \\
                                 & \textbf{Bn-En} & \textbf{En-Bn} & \textbf{Sw-En} & \textbf{En-Sw} & \textbf{Ja-En} & \textbf{En-Ja} & \textbf{De-En} & \textbf{En-De} & \textbf{X-En}     & \textbf{En-X}    \\ \midrule
\textbf{NLLB-200-1.3B}           & 37.78          & 32.83          & 42.66          & 36.28          & 29.60          & 19.07          & 46.23          & 39.91          & 37.51             & 31.00            \\ \midrule
\multicolumn{11}{c}{\textit{\textbf{MetaMath-7B}}}                                                                                                                                                              \\ \midrule
\textbf{MetaMath-7B}             & 1.46           & 0.67           & 3.33           & 1.75           & \textbf{27.62} & 16.76          & 34.36          & 19.42          & 18.62             & 11.92            \\
\textbf{MindMerger}              & 30.76          & -              & 39.43          & -              & 22.50          & -              & 40.05          & -              & 31.57             & -                \\
\textbf{LayAlign}              & 30.91          & -              & 39.02          & -              & 22.36          & -              & 39.43          & -              & 31.98             & -                \\
\textbf{XBridge (Ours)}          & \textbf{35.47} & \textbf{29.23} & \textbf{42.02} & \textbf{34.28} & 24.52          & \textbf{19.60} & \textbf{41.42} & \textbf{35.39} & \textbf{33.37}    & \textbf{29.80}   \\ \midrule
\multicolumn{11}{c}{\textit{\textbf{LLaMA3-8B}}}                                                                                                                                                                \\ \midrule
\textbf{LLaMA3-8B}               & 29.83          & 13.18          & 35.87          & 19.31          & \textbf{27.71} & \textbf{25.40} & 45.28          & \textbf{36.24} & 35.19             & 27.36            \\
\textbf{MindMerger}              & 33.86          & -              & 41.81          & -              & 25.48          & -              & 42.52          & -              & 33.88             & -                \\
\textbf{LayAlign}              & 32.95          & -              & 41.35          & -              & 24.62          & -              & 41.29          & -              & 33.18             & -                \\
\textbf{XBridge (Ours)}          & \textbf{37.09} & \textbf{28.42} & \textbf{44.73} & \textbf{34.68} & 27.63          & 20.12          & \textbf{45.75} & 35.45          & \textbf{36.21}    & \textbf{29.82}   \\ \midrule
\multicolumn{11}{c}{\textit{\textbf{Aya-23-8B}}}                                                                                                                                                                \\ \midrule
\textbf{Aya-23-8B}               & 8.59           & 2.43           & 7.89           & 1.16           & \textbf{29.11} & \textbf{29.34} & \textbf{45.46} & \textbf{38.03} & 28.13             & 23.71            \\
\textbf{MindMerger}              & 33.41          & -              & 41.56          & -              & 24.96          & -              & 41.78          & -              & 33.44             & -                \\
\textbf{LayAlign}              & 32.42          & -              & 40.22          & -              & 24.16          & -              & 41.44          & -              & 32.92             & -                \\
\textbf{XBridge (Ours)}          & \textbf{34.67} & \textbf{28.00} & \textbf{42.88} & \textbf{34.25} & 26.35          & 19.14          & 44.40          & 33.78          & \textbf{33.70}    & \textbf{28.85}   \\ \midrule
\multicolumn{11}{c}{\textit{\textbf{Qwen2.5-7B-Instruct}}}                                                                                                                                                      \\ \midrule
\textbf{Qwen2.5-7B-Instruct}     & 22.15          & 8.30           & 15.05          & 4.35           & 25.92          & \textbf{25.76} & 42.32          & 32.10          & 30.21             & 24.75            \\
\textbf{MindMerger}              & 34.20          & -              & 42.75          & -              & 25.43          & -              & 43.46          & -              & \textbf{34.71}             & -                \\
\textbf{LayAlign}              & 33.39          & -              & 41.26          & -              & \textbf{26.11} & -              & 42.12          & -              & 34.02             & -                \\
\textbf{XBridge (Ours)}          & \textbf{35.89} & \textbf{27.59} & \textbf{43.24} & \textbf{34.55} & 25.50          & 18.66          & \textbf{44.55} & \textbf{33.02} & 34.69    & \textbf{28.64}   \\  \bottomrule
\end{tabular}
\caption{FLORES-101 translation results for stage~1. For clarity, we report results on two low-resource languages (Bengali, Swahili) and two high-resource languages (Japanese, German), with complete results and COMET scores in Appendix~\ref{detailed_results}. "X" denotes all languages except for English. We bold the best scores for each LLM group.}
\label{table_main_flores}
\end{table*}

\subsection{Experiment Setup}

\paragraph{Base Models}
We evaluate XBridge on four representative base LLMs: MetaMath-7B-V1.0~\cite{yu2024metamath}, LLaMA3-8B~\cite{grattafiori2024llama}, Aya-23-8B~\cite{ustun2024aya}, and Qwen2.5-7B-Instruct~\cite{qwen2025qwen25technicalreport}.
As the pretrained encoder-decoder NMT model, we adopt NLLB-200-1.3B~\cite{nllbteam2022languageleftbehindscaling}, which covers 200 languages with strong multilingual capacity.

\paragraph{Baselines}
We compare XBridge with these strong baselines:
(1) \textbf{SFT} performs multilingual instruction fine-tuning directly on each base LLM.
(2) \textbf{Translate-Test}~\cite{artetxe-etal-2023-revisiting} translates inputs to English, queries the English-SFT LLM, and translates the output back to the target language.
(3) \textbf{MindMerger}~\cite{huang2024mindmerger} augments the LLM input with a pretrained multilingual encoder to enhance multilingual understanding, forming a strong multilingual-to-English system.
(4) \textbf{LayAlign}~\cite{ruan-etal-2025-layalign} further extends MindMerger with layer-wise fusion strategies to better integrate encoder representations into the LLM.

\paragraph{Language Setup}
\label{language_text}
Following \citet{chen2024breaking}, we experiment on ten languages: Bengali (Bn), German (De), English (En), Spanish (Es), French (Fr), Japanese (Ja), Russian (Ru), Swahili (Sw), Thai (Th), and Chinese (Zh).
These languages span diverse language families and resource levels.
We treat Bn, Sw, and Th as low-resource languages, and the remaining as high-resource ones.

\paragraph{Training Datasets}
For stage~1 training, we extract English-centric translation pairs from OPUS-100~\cite{zhang-etal-2020-improving}.
For XBridge, we further translate the English sentences into other languages $L_y$ using NLLB-200-3.3B, constructing trilingual \textit{x-en-y} data.
For stage~2 and stage~3, we adopt multilingual mathematical reasoning data from \citet{ruan-etal-2025-layalign} and multilingual abstractive summarization data from XL-Sum~\cite{hasan-etal-2021-xl}. For XBridge, we construct bilingual responses using NLLB-200-3.3B. 
Appendix~\ref{details_training} presents details about data processing and statistics.

\begin{figure*}[t]
    \begin{subfigure}{\textwidth}
    \centering
    \includegraphics[width=\linewidth]{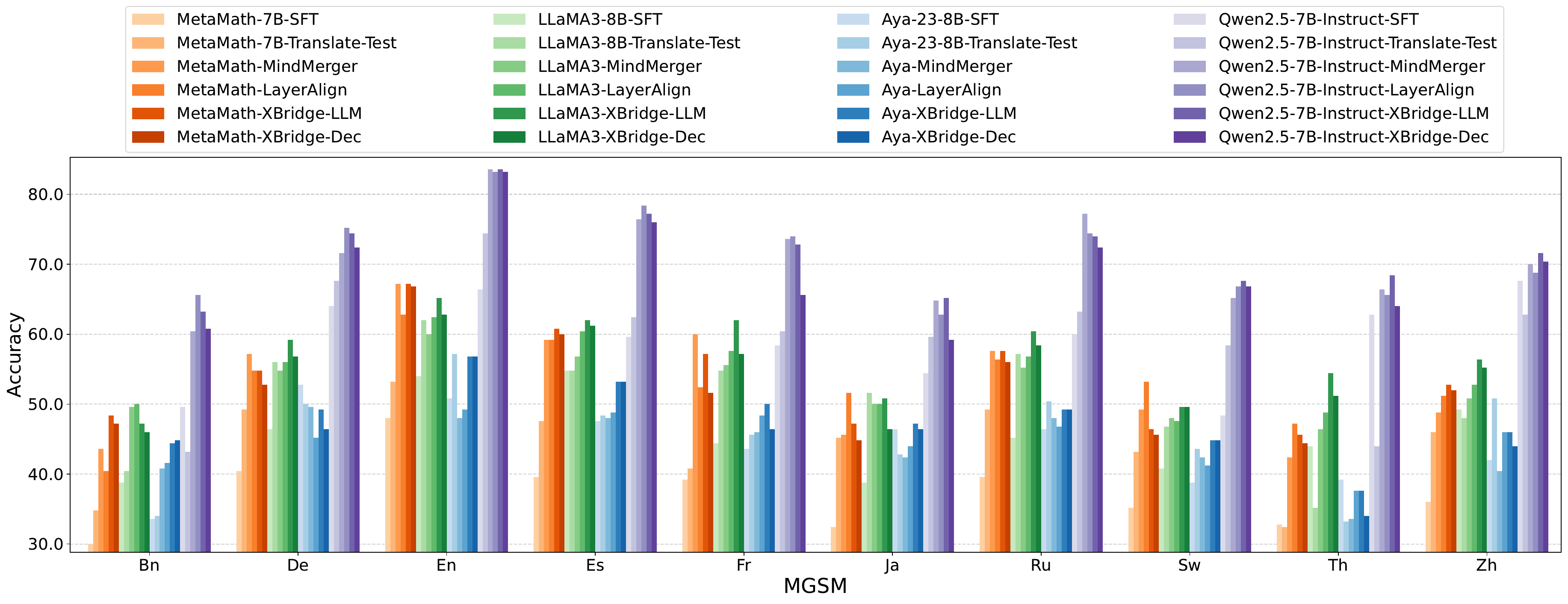}
    \end{subfigure}
    \hfill
    \begin{subfigure}{\textwidth}
    \centering
    \includegraphics[width=\linewidth]{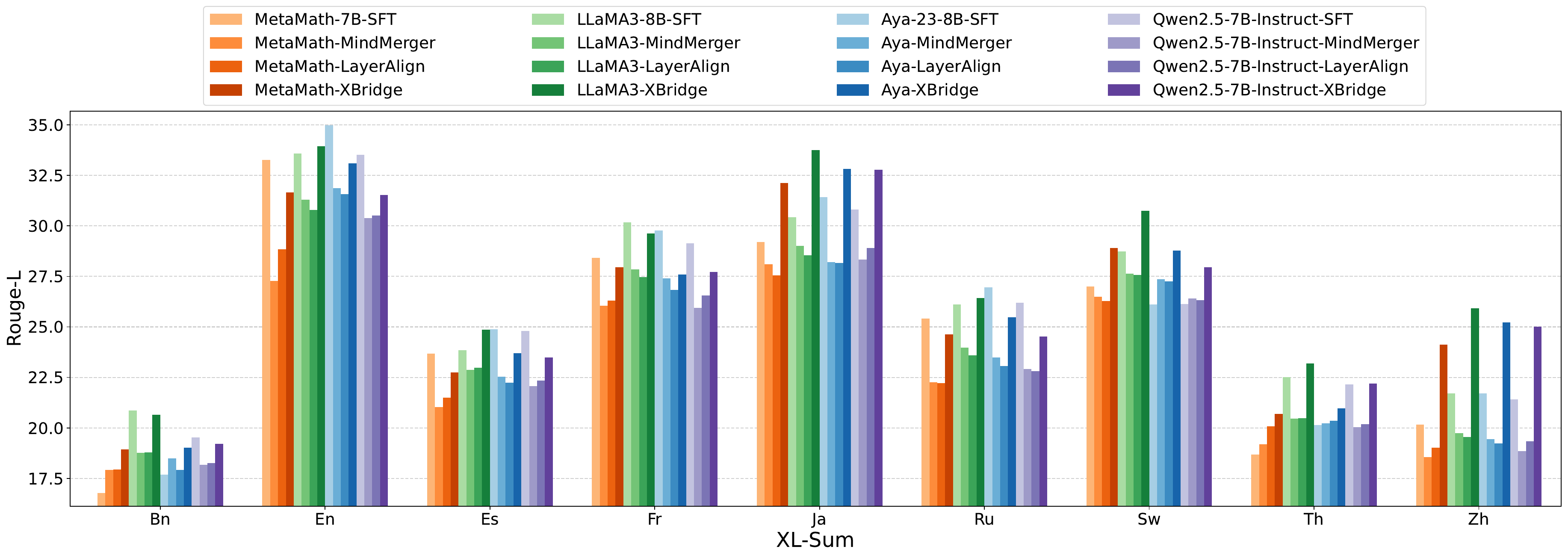}
    \end{subfigure}
    \caption{Multilingual reasoning accuracy on MGSM and multilingual summarization Rouge-L on XL-Sum, with complete results in Appendix~\ref{detailed_results}. Models with the same base LLM share the same color scheme, where lighter shades denote baselines and darker shades denote XBridge. "XBridge-LLM" refers to English reasoning by the LLM, while "XBridge-Dec" refers to multilingual reasoning by the composed decoder. For XL-Sum, since the baselines produce English-only summaries, we translate them into target languages using NLLB-200-1.3B for evaluation.}
    \label{table_main_mgsm_xlsum}
\end{figure*}

\paragraph{Evaluation Benchmarks}
For stage~1, we evaluate cross-model mapping quality on FLORES-101~\cite{goyal-etal-2022-flores}.
Given the strong English ability of LLMs, we use \textit{x-en} and \textit{en-x} translation performance to measure multilingual understanding and generation, respectively, and report BLEU~\cite{PapineniRWZ02} and COMET~\cite{rei-etal-2020-comet} scores. 
For base LLMs, we leverage \textit{MMT-LLM}~\cite{zhu-etal-2024-multilingual} framework to evaluate translation capability in a 1-shot setting.
For stage~2 and stage~3, we evaluate multilingual reasoning on MGSM~\cite{shilanguage} with Accuracy, and multilingual abstractive summarization on XL-Sum with multilingual Rouge-L~\cite{lin2004rouge}.

\paragraph{Model Configuration and Training Details}
The encoder-side mapping is implemented as a two-layer multi-layer perceptron (MLP), while the decoder-side mapping is a four-layer MLP composed of two stacked two-layer MLP blocks.
All intermediate dimensions are aligned with the LLM hidden size.
We use the AdamW optimizer with a learning rate of $2\times10^{-5}$, train each stage for 3 epochs with a batch size of 128, and conduct experiments on 8 NVIDIA H800 GPUs.
We empirically set $\lambda_1 = 1.0$, $\lambda_2 =1.0$, and $\lambda_2 = 6.0$ when the corresponding losses are active, with detailed activation schedules described in Section~\ref{subsection_training}.

\subsection{Experimental Results}

\paragraph{XBridge effectively offloads multilingual capability to the external multilingual model, while preserving the LLM as a knowledge and reasoning core.}

Table~\ref{table_main_flores} evaluates the cross-model mapping learned in stage~1 on FLORES-101.
Across all base LLMs, XBridge substantially improves both multilingual understanding and generation, with especially large gains on low-resource languages where base LLMs have limited capability.
The performance of XBridge approaches that of the external NLLB-200-1.3B and outperforms encoder-augmented baselines, showing that XBridge can effectively offload multilingual ability to external NMT models while keeping the LLM frozen as a knowledge and reasoning core.
Importantly, performance on high-resource languages remains comparable to base LLMs, indicating that offloading does not degrade the original strengths of LLMs.

\paragraph{Encoder adaptation improves multilingual understanding without degrading English performance.}

Figure~\ref{table_main_mgsm_xlsum} presents multilingual reasoning accuracy on MGSM after encoder adaptation.
XBridge outperforms the base LLM, encoder-only baselines, and the Translate-Test pipeline.
Since MGSM accuracy is language-agnostic, these gains directly reflect better semantic transfer between multilingual encoder representations and the LLM reasoning space.
These results indicate that encoder-side adaptation facilitates more effective utilization of multilingual representations by the LLM, improving multilingual reasoning without sacrificing its English-centric reasoning capability.

\begin{figure*}[t]
\centering
\begin{subfigure}{0.318\textwidth}
\centering
\includegraphics[width=\linewidth]{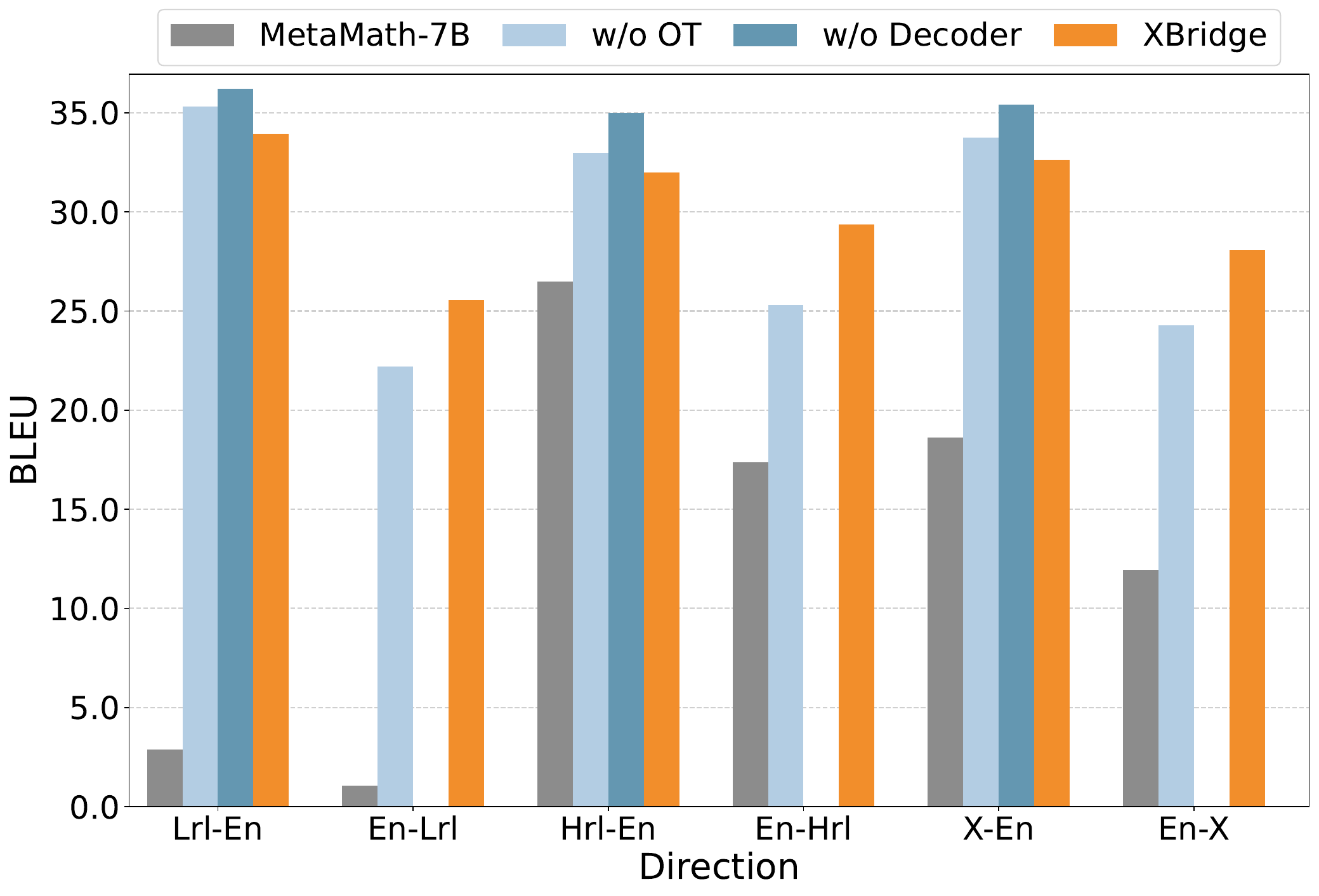}
\caption{FLORES-101}
\end{subfigure}
\begin{subfigure}{0.458\textwidth}
\centering
\includegraphics[width=\linewidth]{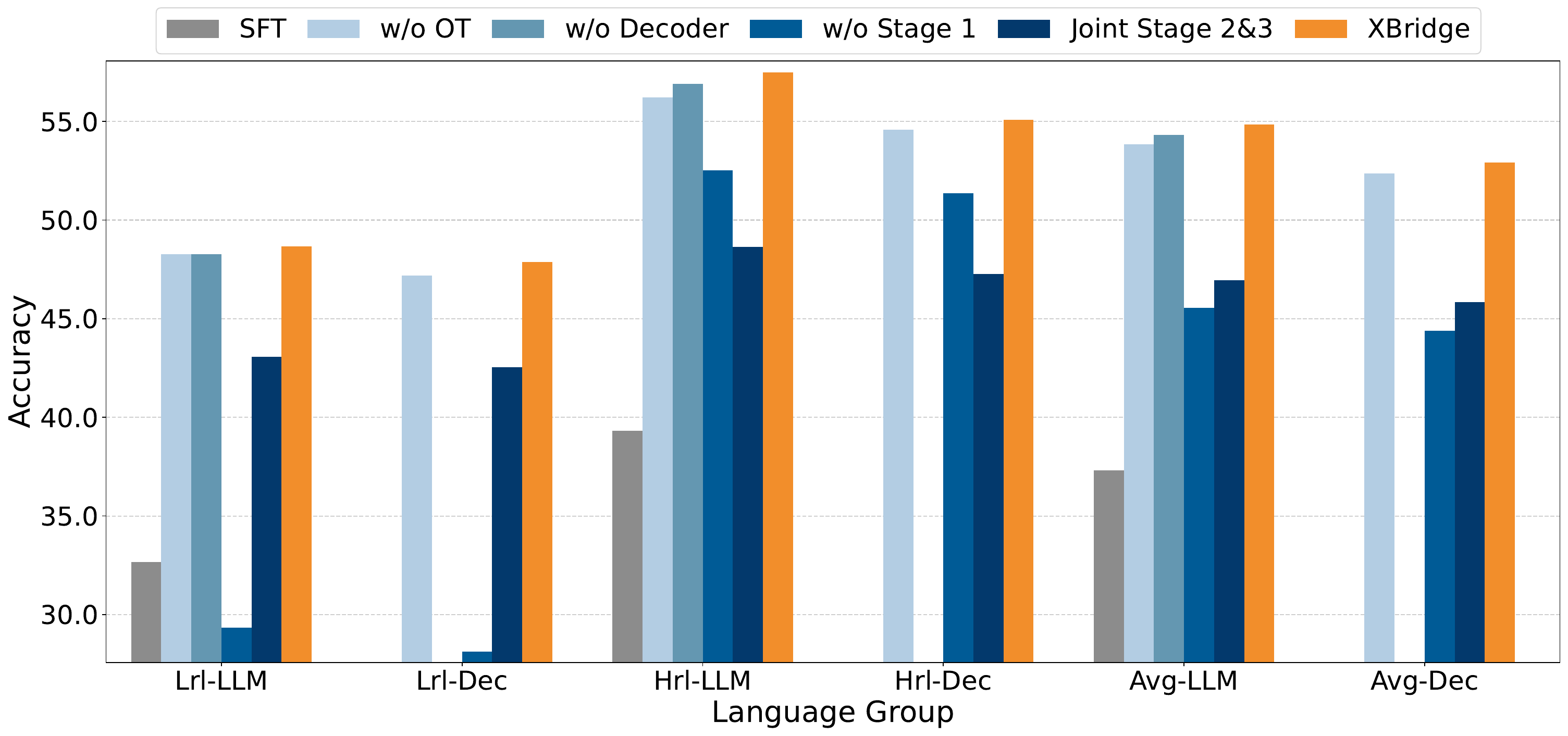}
\caption{MGSM}
\end{subfigure}
\begin{subfigure}{0.21\textwidth}
\centering
\includegraphics[width=\linewidth]{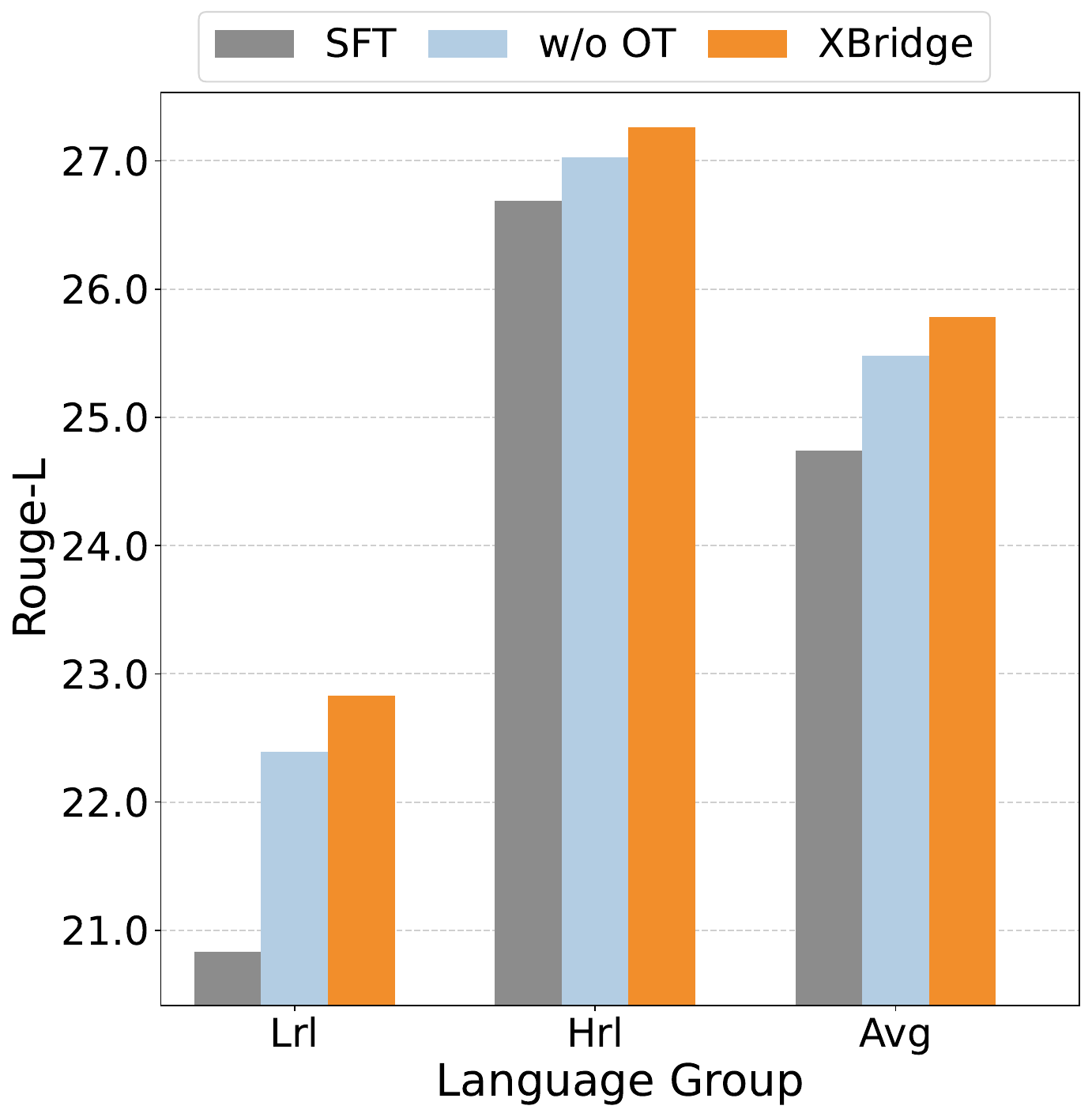}
\caption{XL-Sum}
\end{subfigure}
\caption{Ablation analysis of XBridge. We compare different ablated variants of XBridge: encoder-only augmentation "w/o Decoder", loss ablation "w/o OT", removal of stage~1 "w/o Stage~1", and joint optimization of stage~2\&3 "Joint Stage~2\&3". “Lrl”, “Hrl”, and “Avg” denote low-, high-resource, and average performance, respectively.}
\label{ablation}
\end{figure*}

\paragraph{Decoder adaptation achieves faithful multilingual generation.}

We further evaluate decoder adaptation on MGSM and XL-Sum in Figure~\ref{table_main_mgsm_xlsum}.
On MGSM, decoder-generated multilingual reasoning (XBridge\_Dec) achieves accuracy comparable to English LLM outputs, suggesting that the decoder can faithfully express reasoning content across languages.
On XL-Sum, XBridge consistently outperforms encoder-augmented baselines and achieves better average performance than the SFT baseline, with particularly clear gains on languages where multilingual generation is more challenging.
While translation-cascaded systems are limited by the NMT model, XBridge directly leverages the LLM’s knowledge through decoder adaptation, resulting in more stable multilingual generation across languages.
These results demonstrate the importance of decoder adaptation for robust multilingual generation.

\section{Analysis}

\subsection{Ablation Analysis}
\label{ablation_text}


\begin{figure*}[t]
\centering
  \includegraphics[width=1.0\linewidth]{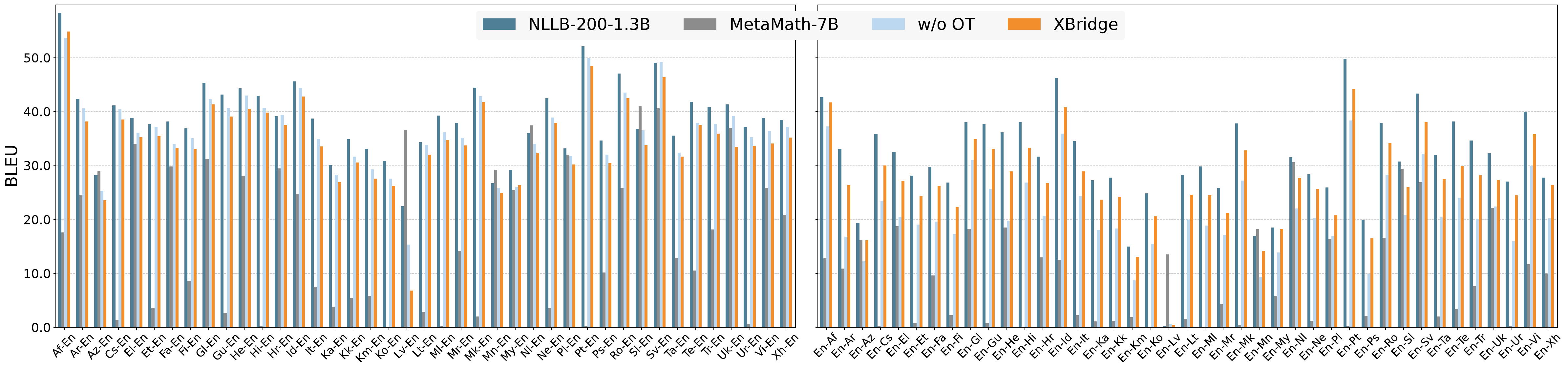}
  \caption{Cross-lingual generalization to 41 untuned languages in FLORES-101. Left: \textit{X$\rightarrow$En} direction. Right: \textit{En$\rightarrow$X} direction. We directly evaluate the ablation variants described in Section~\ref{ablation_text}. Appendix~\ref{detailed_results} lists the included untuned languages and provides detailed results.}
\label{figure_untuned_languages}
\end{figure*}

We conduct the ablation study on MetaMath-7B-V1.0 to analyze the contribution of each component and training strategies in XBridge, and evaluate ablated variants on FLORES-101, MGSM, and XL-Sum. Figure~\ref{ablation} presents the results, and Appendix~\ref{detailed_results} provides detailed results.

\paragraph{Encoder-Decoder Collaboration}
Removing the decoder (\textit{w/o Decoder}) achieves competitive multilingual-to-English understanding but fails to support multilingual generation, and underperforms XBridge on MGSM.
This confirms that encoder-only augmentation is insufficient for multilingual reasoning and generation.

\paragraph{OT Alignment Objectives}
Similarly, removing the OT alignment (\textit{w/o OT}) leads to performance degradation on all benchmarks, particularly for multilingual generation, indicating that token-level soft alignment plays a crucial role in bridging heterogeneous representation spaces between the LLM and the multilingual decoder.

\paragraph{Stage-Wise Optimization}
Skipping stage~1 (\textit{w/o Stage~1}) results in a substantial performance drop across all metrics, suggesting that direct task-level adaptation is insufficient when the representation gap between the LLM and the multilingual model remains large.
Moreover, jointly training stage~2 and stage~3 (\textit{Joint Stage~2\&3}) underperforms the stage-wise optimization, reflecting a trade-off between LLM- and decoder-side generation objectives.
These results support the design of stage-wise adaptation, where coarse-grained cross-model alignment is first established, followed by fine-grained encoder and decoder specialization, enabling XBridge to achieve stable and effective multilingual reasoning and generation.

\begin{figure}[t]
  \centering
  \begin{subfigure}[t]{0.494\columnwidth}
    \centering
    \includegraphics[width=\linewidth]{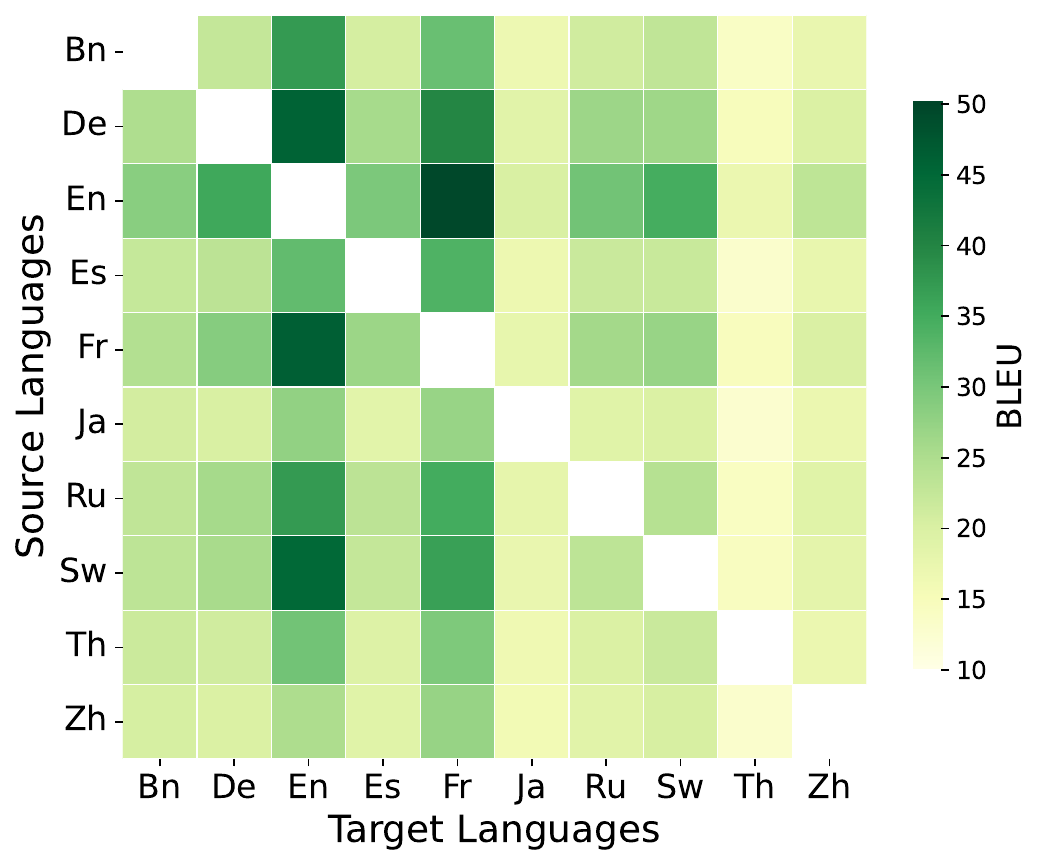}
    \caption{FLORES-101}
  \end{subfigure}
  \begin{subfigure}[t]{0.494\columnwidth}
    \centering
    \includegraphics[width=\linewidth]{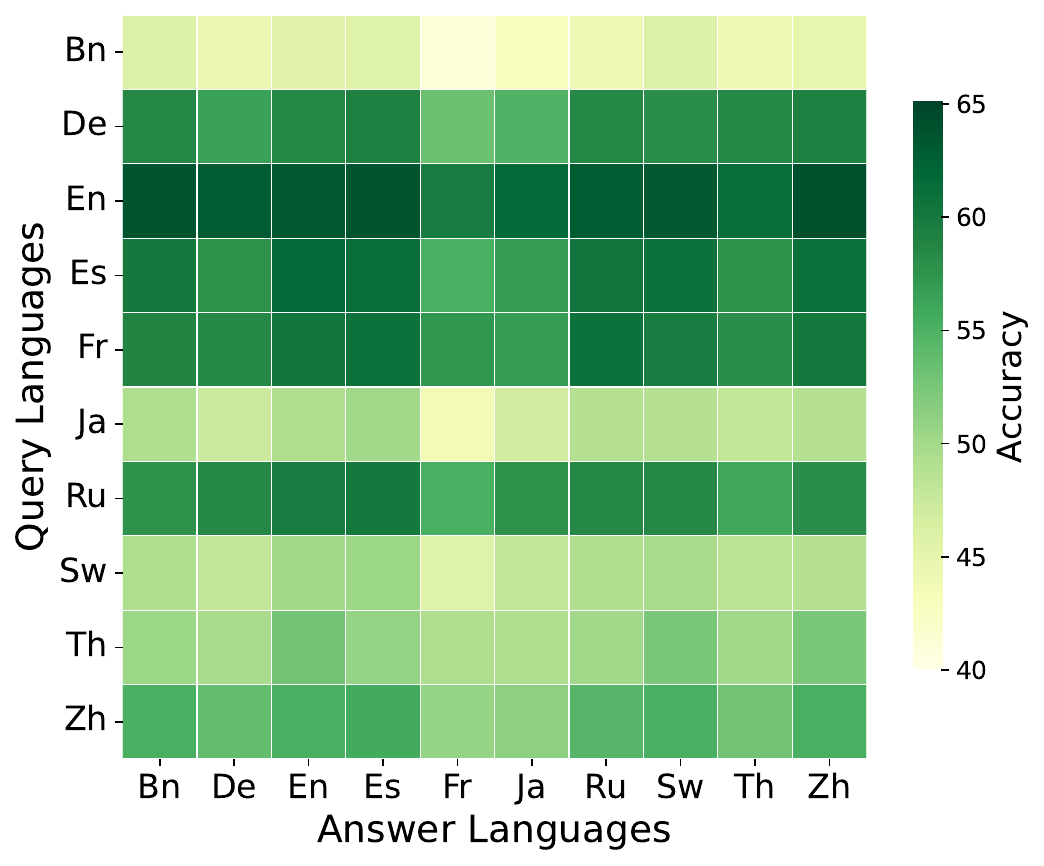}
    \caption{MGSM}
  \end{subfigure}
  \caption{Evaluation for language-on-demand generation. Appendix~\ref{detailed_results} presents detailed results.}
  \label{heatmap}
\end{figure}

\subsection{Generalization to Untuned Languages}
\label{untuned_language_text}

To examine whether the cross-model mappings learned by XBridge are language-agnostic rather than simply tied to specific training languages, we evaluate cross-model cross-lingual transfer on 41 untuned languages (listed in Table~\ref{language_statistics}) in Figure~\ref{figure_untuned_languages}, based on variants of Section~\ref{ablation_text}.

XBridge yields substantial gains on untuned languages over the base LLM, with performance approaching the external NLLB model.
This indicates that stage~1 cross-model mapping learns language-agnostic semantic transfer that generalizes beyond tuned languages, rather than language-specific mapping.
Meanwhile, performance in the \textit{En$\rightarrow$X} directions highlights the importance of optimal transport.
Removing the OT objective leads to a substantial drop in generation quality, particularly where tokenization length differs across different tokenizers.
These results suggest that OT enables robust alignment between heterogeneous tokenizations, which is crucial for generalizable multilingual generation.
Overall, the results demonstrate that cross-model semantic alignment generalizes across languages, while OT is crucial for achieving reliable generation-level generalization.

\begin{figure}[t]
  \centering
  \begin{subfigure}[t]{0.49\columnwidth}
    \centering
    \includegraphics[width=\linewidth]{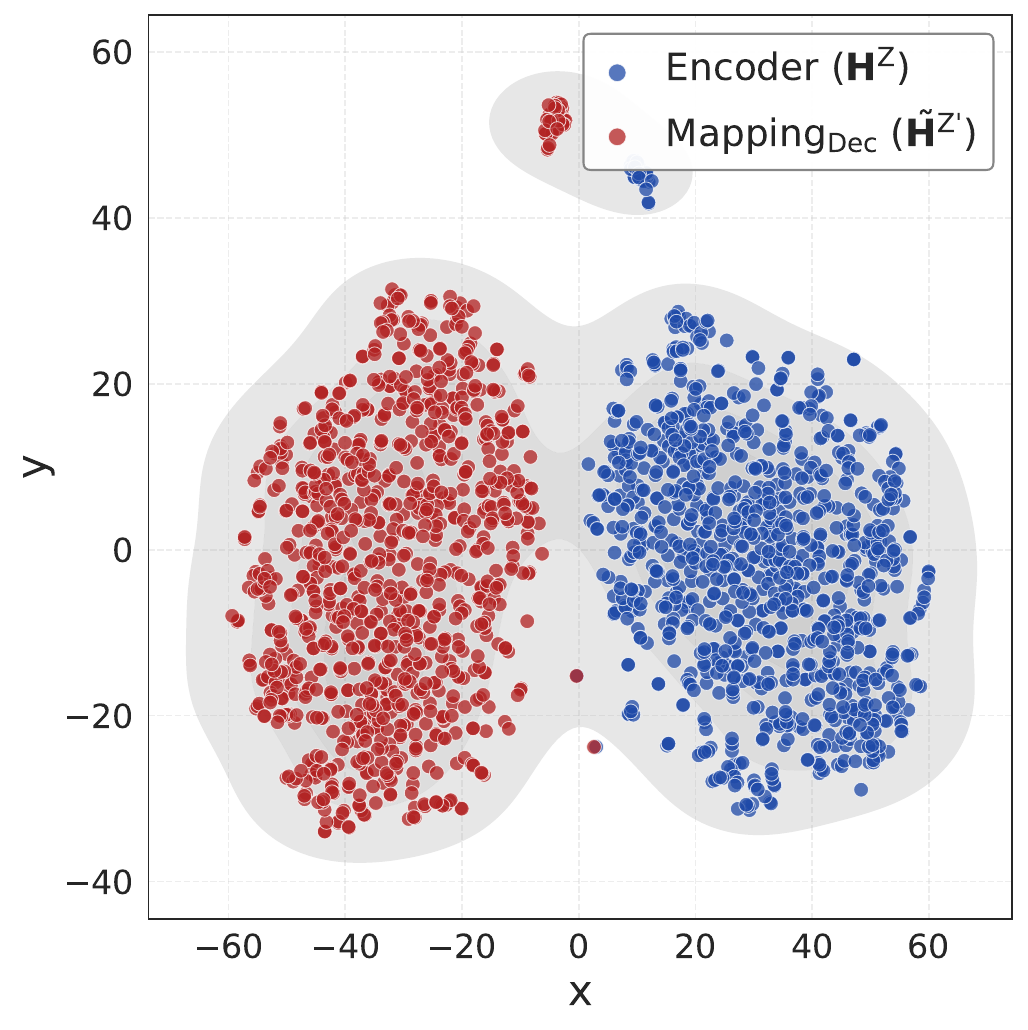}
    \caption{w/o OT}
  \end{subfigure}
  \begin{subfigure}[t]{0.49\columnwidth}
    \centering
    \includegraphics[width=\linewidth]{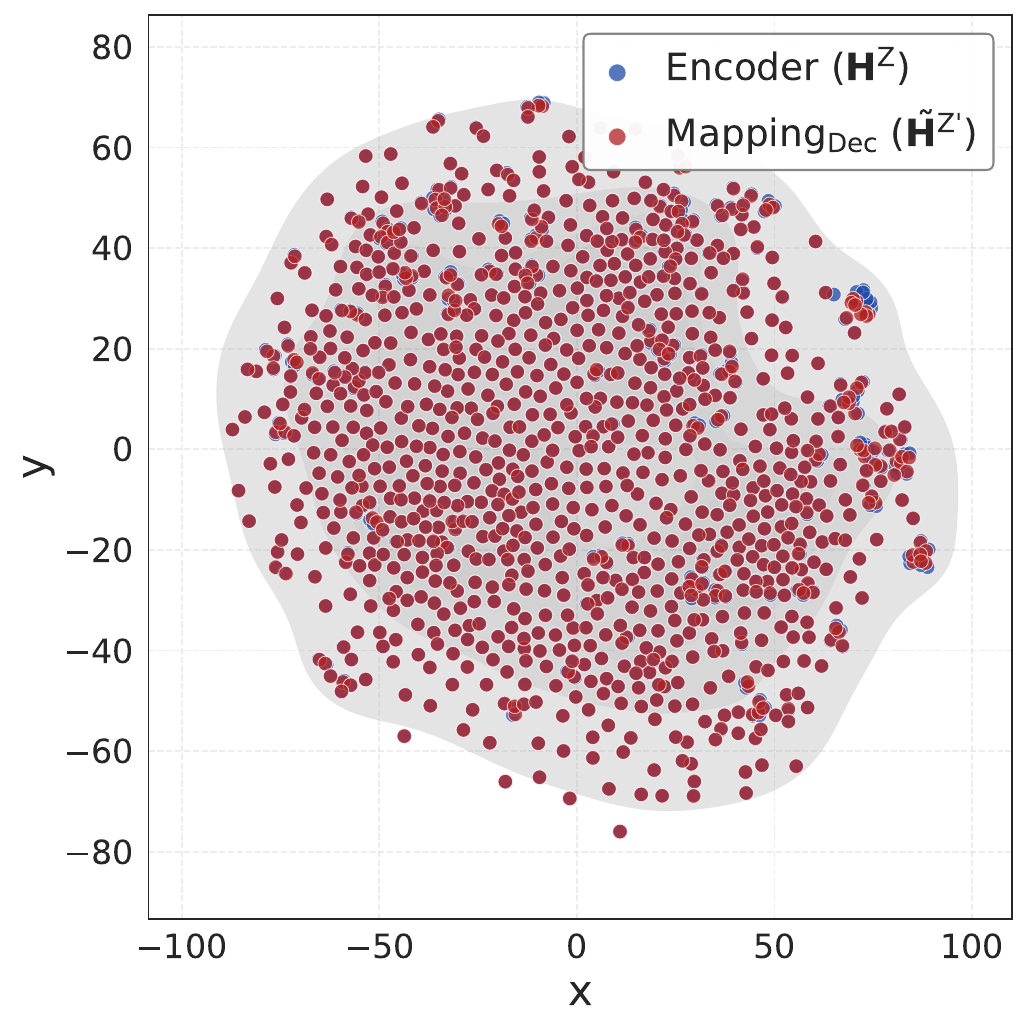}
    \caption{w/ OT}
  \end{subfigure}
  \caption{Visualization of sentence-level representation alignment for Chinese (Zh). We compare models trained without OT (left) and with OT (right) using t-SNE.}
  \label{visualization}
\end{figure}

\subsection{Language-on-Demand Generation}
\label{cross_lingual_text}

We verify the language-on-demand property of XBridge by switching the target language token $\langle y \rangle$ to generate outputs in arbitrary languages without retraining in Figure~\ref{heatmap}.

On FLORES-101, we evaluate translation between all language pairs. With the target language fixed, changing the source language causes only minor performance differences, while variations primarily depend on the target language.
On MGSM, we force the decoder to generate responses in languages different from the input query language. For each input language, performance remains largely stable across different output languages.
These results indicate that XBridge enables stable language-on-demand generation, supporting flexible multilingual outputs while preserving a language-agnostic reasoning core in the LLM.

\begin{figure}[t]
  \centering
  \begin{subfigure}[t]{0.45\columnwidth}
    \centering
    \includegraphics[width=\linewidth]{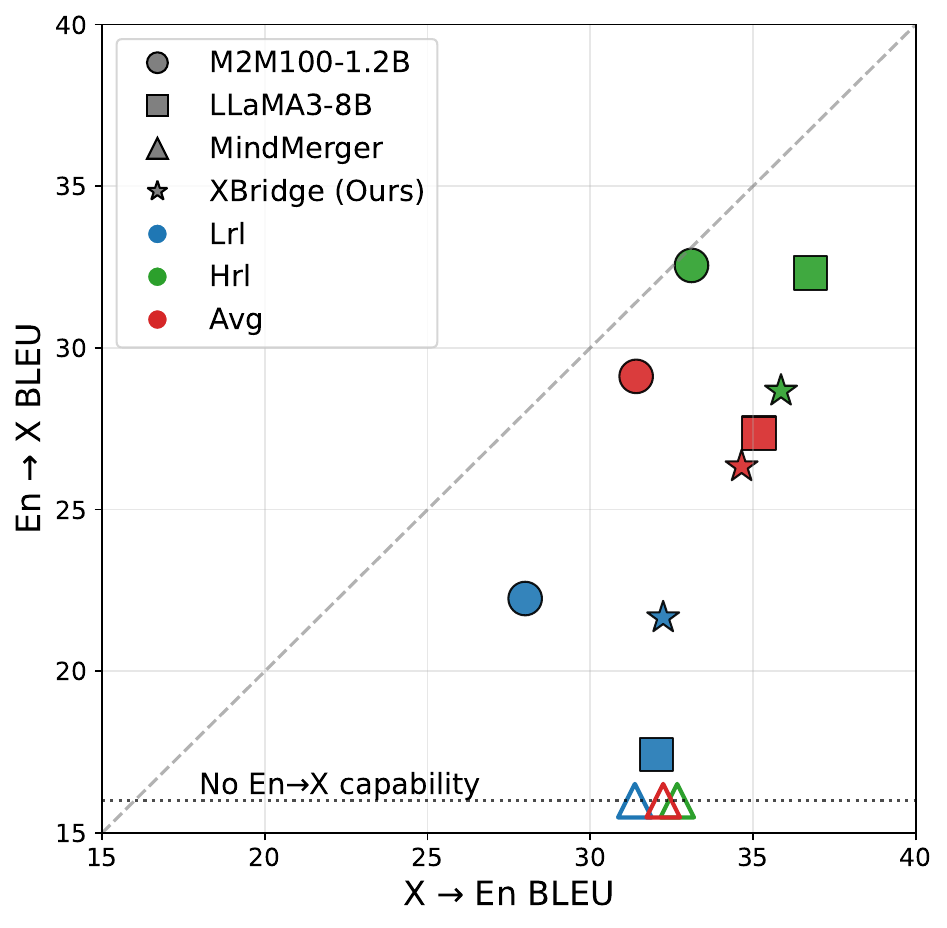}
    \caption{FLORES-101}
  \end{subfigure}
  \begin{subfigure}[t]{0.53\columnwidth}
    \centering
    \includegraphics[width=\linewidth]{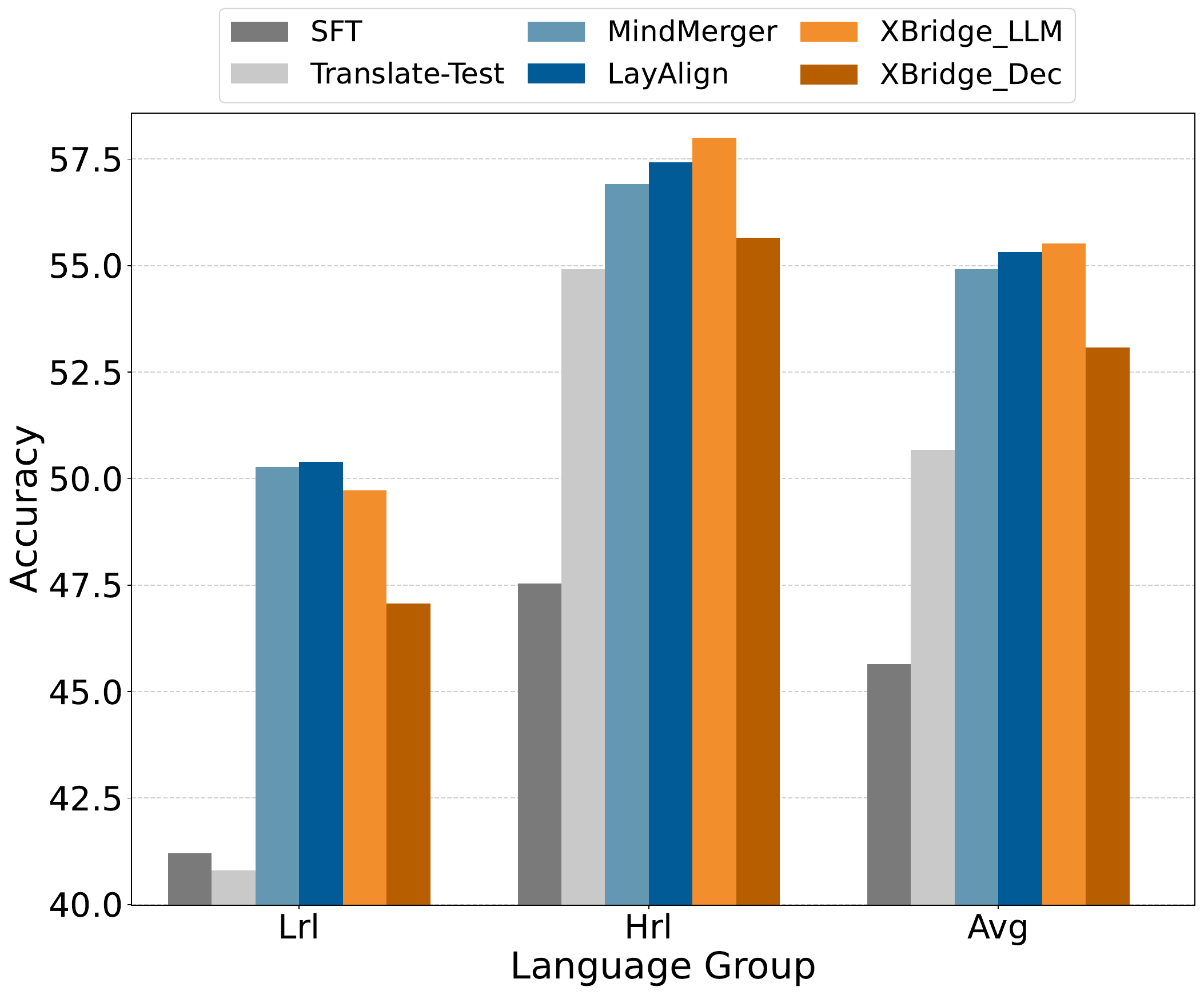}
    \caption{MGSM}
  \end{subfigure}
  \caption{LLaMA3-8B composed with M2M100-1.2B. “Lrl”, “Hrl”, and “Avg” denote low-resource, high-resource, and average performance, respectively. Hollow markers placed on the bottom boundary indicate models that lack \textit{En$\rightarrow$X} translation capability. Appendix~\ref{detailed_results} presents detailed results.}
  \label{m2m100}
\end{figure}

\subsection{Representation Visualization}

To analyze the effect of optimal transport (OT) on aligning heterogeneous representations, we visualize sentence-level hidden states for each language.
Specifically, we compare encoder representations of LLM English outputs $\mathbf{H}^\mathbf{z}$ with decoder-side representations after mapping $\tilde{\mathbf{H}}^{\mathbf{z}'}$. We obtain sentence-level vectors via average pooling and project them to 2-dim for visualization using t-SNE~\cite{van2008visualizing}.

As shown in Figure~\ref{visualization}, without OT, the two sets of representations form largely separate clusters, reflecting a substantial distribution gap at the LLM-decoder interface. 
In contrast, when OT is applied, the two sets of representations overlap substantially, with density contours largely merged, indicating that OT promotes fine-grained semantic consistency and reduces token-level misalignment across model components.

\subsection{Composing with Different NMT Models}
\label{m2m100_text}

To further examine the generality of XBridge beyond a specific NMT backbone, we replace the multilingual NMT model with M2M100-1.2B~\cite{fan2021beyond} in Figure~\ref{m2m100} while keeping the same training and evaluation settings as in Section~\ref{main_experiment_text}. 

XBridge remains effective when composed with M2M100-1.2B. On FLORES-101, XBridge achieves strong cross-model transfer across low- and high-resource language directions, demonstrating that the lightweight mapping layers can reliably bridge NMT models and LLMs. On MGSM, XBridge outperforms the translation-cascaded baseline, indicating that the benefits of XBridge extend beyond translation quality to multilingual reasoning. Overall, these results demonstrate that XBridge is an architecture-agnostic framework that generalizes across both different LLM backbones and multilingual NMT backbones.

\begin{figure}[t]
  \centering
  \begin{subfigure}[t]{0.45\columnwidth}
    \centering
    \includegraphics[width=\linewidth]{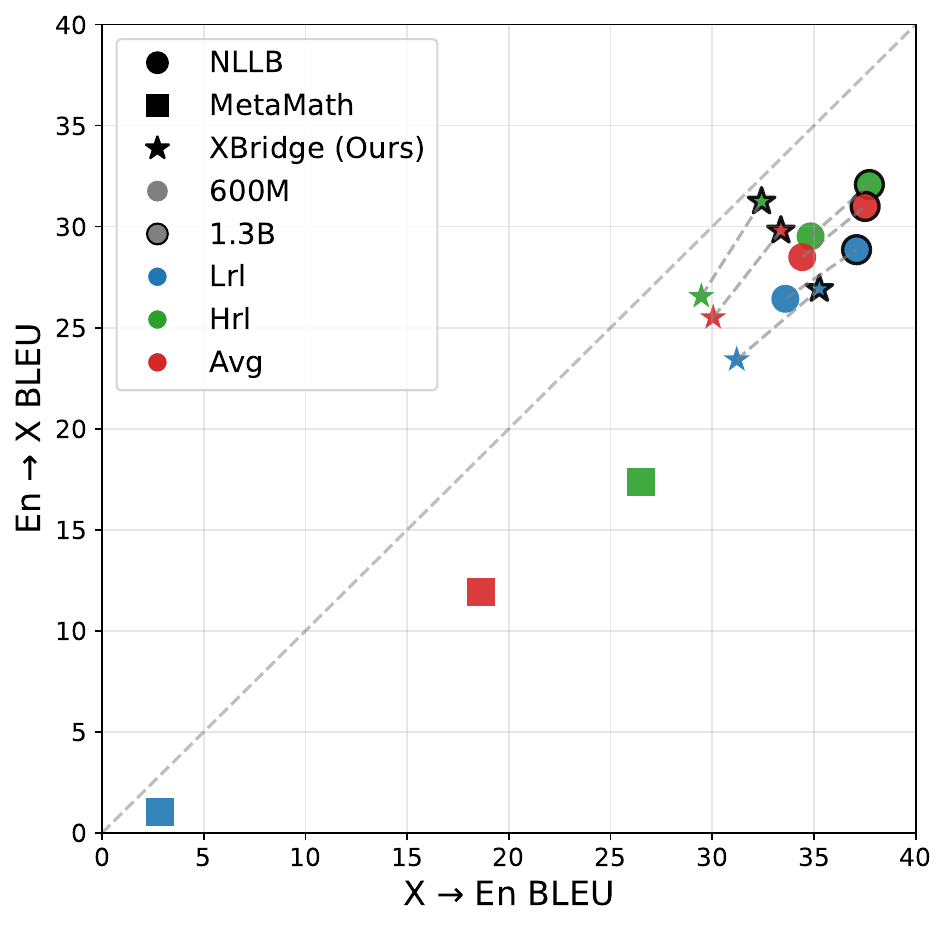}
    \caption{FLORES-101}
  \end{subfigure}
  \begin{subfigure}[t]{0.53\columnwidth}
    \centering
    \includegraphics[width=\linewidth]{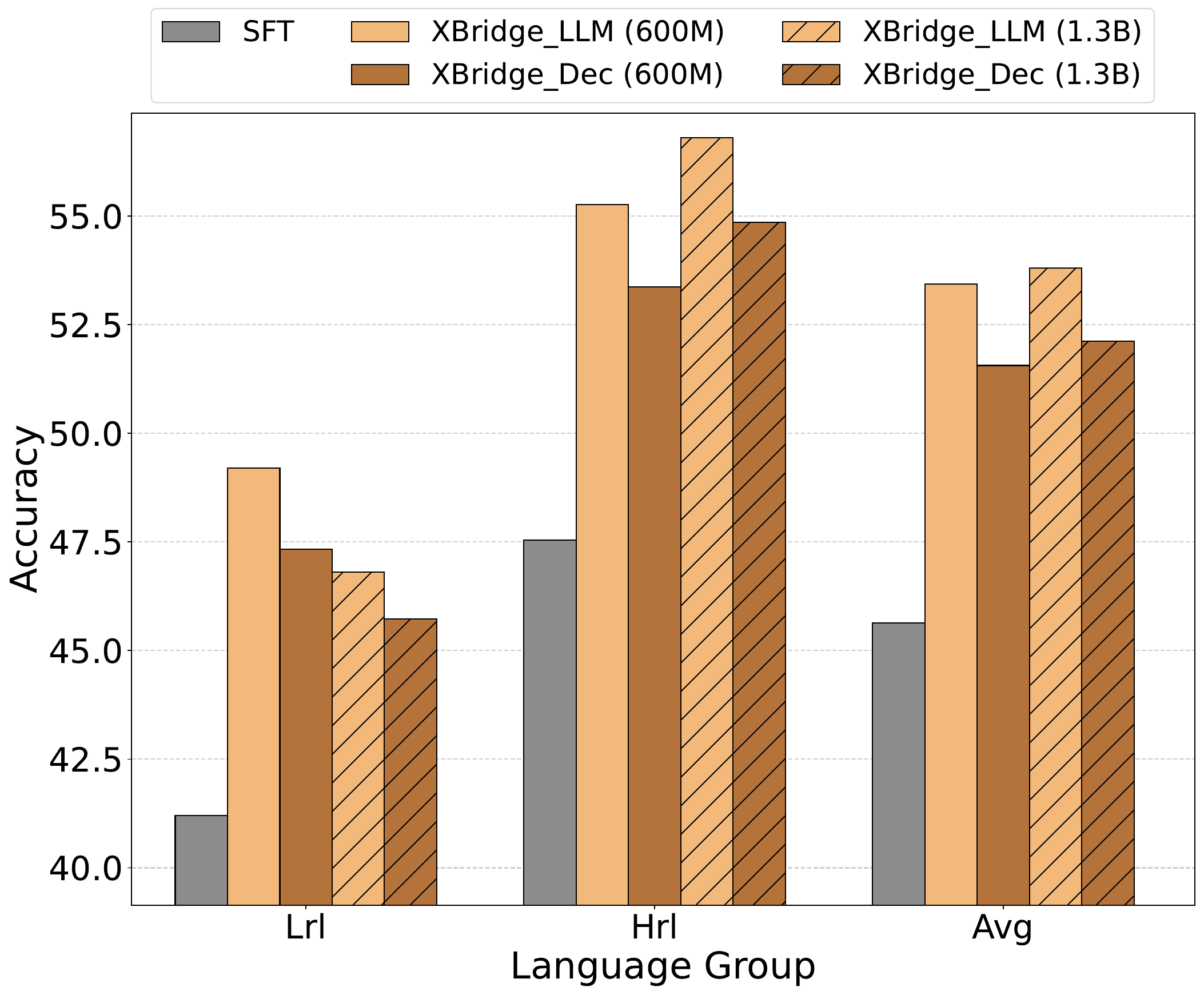}
    \caption{MGSM}
  \end{subfigure}
  \caption{MetaMath-7B composed with NLLB in different sizes (600M vs. 1.3B). “Lrl”, “Hrl”, and “Avg” denote low-, high-resource, and average performance, respectively. Appendix~\ref{detailed_results} presents detailed results.}
  \label{different_nmt_size}
\end{figure}

\subsection{Impact of NMT Model Size}
\label{nmt_size_text}

We further investigate the impact of NMT model capacity on XBridge by comparing NLLB-200-600M and NLLB-200-1.3B, both integrated with MetaMath-7B. Figure~\ref{different_nmt_size} presents the results.

On FLORES-101, larger NLLB models consistently improve multilingual capability across languages, indicating that stronger multilingual capacity in the composed NMT model leads to better multilingual understanding and generation.
On MGSM, increasing the NLLB size brings only marginal changes in reasoning accuracy, suggesting that reasoning performance is primarily determined by the LLM core. 
These results align with our design, indicating that the quality of the composed NMT model directly influences cross-model mapping, while reasoning remains governed by the LLM.

\subsection{Supplementary Analysis}

We conduct supplementary analysis, including efficiency analysis (Appendix~\ref{efficiency_text}), the case study on MGSM (Appendix~\ref{case_study_text}), and evaluation on multilingual commonsense reasoning (Appendix~\ref{csqa_text}).
\section{Conclusion}

In this paper, we propose XBridge, a compositional framework that offloads multilingual capability to an external encoder-decoder NMT model, while preserving the LLM as an English-centric core for general knowledge processing. 
Extensive experiments demonstrate that XBridge enables efficient multilingual extension, raising low-resource and unseen language performance to near external NMT, without compromising LLM's core abilities.
\section*{Limitations}

While XBridge substantially mitigates multilingual imbalance, notably improving performance on low-resource and previously unseen languages for LLMs, the overall model still exhibits some imbalance in multilingual capabilities. 
This is primarily due to the combined influence of the external encoder-decoder NMT model and the base LLM, which limits complete uniformity across languages. 
Future work could further explore strategies to harmonize these components.

\section*{Acknowledgements}

We thank all the anonymous reviewers for their insightful and valuable comments on this paper. This work was supported by the grant from the Beijing Natural Science Foundation (No. L257006).

\bibliography{custom}

\newpage
\appendix

\section{Optimal Transport Algorithm for Heterogeneous Representations}
\label{ot_algorithm}

In this section, we briefly review the standard OT formulation and describe how it is adapted to align heterogeneous and unequal-length representation sequences in XBridge.

\subsection{Optimal Transport Between Discrete Distributions}

Optimal Transport (OT) provides a principled framework for measuring the discrepancy between two probability distributions by minimizing the cost of transporting probability mass from one distribution to the other~\cite{peyre2019computational}.
Consider the following discrete transport problem: given two probability distributions $P$ and $Q$, 
\begin{equation}
\begin{aligned}
    P = \{(w_i, m_i)\}_{i=1}^{n}, \quad \text{s.t.} \sum_{i=1}^{n} m_i = 1, \\
    Q = \{(w'_j, m'_j)\}_{j=1}^{n'}, \quad \text{s.t.} \sum_{j=1}^{n'} m_j = 1.
\end{aligned}
\end{equation}
where each support point $w_i, w'_j \in \mathbb{R}^d$ is associated with a non-negative probability mass $m_i, m'_j$.
Given a cost function $c(w_i, w'_j)$ that measures the unit cost of transporting mass from $w_i$ to $w'_j$, the transport cost between $P$ and $Q$ is defined as:
\begin{equation}
\begin{aligned}
\mathcal{D}(P, Q)
&= \min_{\mathbf{T} \ge 0} \sum_{i,j} \mathbf{T}_{ij} \, c(w_i, w'_j), \\
\text{s.t. } &\sum_{j=1}^{n'} \mathbf{T}_{ij} = m_i, \forall i \in \{1,\ldots,n\},\\
&\sum_{i=1}^{n} \mathbf{T}_{ij} = m'_j, \forall j \in \{1,\ldots,n'\}.
\end{aligned}
\end{equation}
where $\mathbf{T}_{ij}$ denotes the mass transported from $w_i$ to $w'_j$.

\subsection{OT for Aligning Unequal-Length Representation Sequences}

In XBridge, we apply OT to align two heterogeneous token representation sequences:
\begin{equation}
\begin{aligned}
    &\mathbf{H}^{\mathbf{z}} = (H^{\mathbf{z}}_1, \ldots, H^{\mathbf{z}}_k), \\
    &\tilde{\mathbf{H}}^{\mathbf{z}'} = (\tilde{H}^{\mathbf{z}'}_1, \ldots, \tilde{H}^{\mathbf{z}'}_m).
\end{aligned}
\end{equation}
where $k \neq m$ in general due to different tokenization schemes.
Both sequences originate from the same underlying LLM output but are obtained through different encoding pathways, making explicit token-wise correspondence unavailable.

We formulate their alignment as the following OT problem:
\begin{equation}
\begin{aligned}
\mathcal{D}(\mathbf{H}^{\mathbf{z}}&, \tilde{\mathbf{H}}^{\mathbf{z}'})
= \min_{\mathbf{T} \ge 0} \sum_{i,j} \mathbf{T}_{ij} \,
c(H^{\mathbf{z}}_i, \tilde{H}^{\mathbf{z}'}_j), \\
\text{s.t. } &\sum_{j=1}^{m} \mathbf{T}_{ij} = m_i^{\mathbf{z}}, \forall i \in \{1,\ldots,k\}, \\
&\sum_{i=1}^{k} \mathbf{T}_{ij} = m_j^{\mathbf{z}'}, \forall j \in \{1,\ldots,m\}.
\end{aligned}
\end{equation}
where the cost function $c(\cdot,\cdot)$ is defined as cosine distance.

The probability masses $m_i^{\mathbf{z}}$ and $m_j^{\mathbf{z}'}$ are obtained by normalizing the $\ell_1$ norms of the corresponding representations.
This choice is motivated by prior work~\citep{schakel2015measuring, yokoi2020word} showing that embedding norms correlate with token importance, with semantically salient words exhibiting larger magnitudes.

\subsection{Approximate OT via Relaxed Marginal Constraints}

Solving the exact OT problem requires $O(n^3)$ linear programming, which is computationally prohibitive for long sequences.
While entropic regularization methods such as Sinkhorn~\cite{cuturi2013sinkhorn} or IPOT~\cite{xie2020fast} provide approximate solutions, they still introduce significant overhead during training.

Following \citet{kusner2015word}, we adopt a relaxed OT formulation by removing the second marginal constraint:
\begin{equation}
\begin{aligned}
\mathcal{D}^*(\mathbf{H}^{\mathbf{z}}, \tilde{\mathbf{H}}^{\mathbf{z}'})
= \min_{\mathbf{T} \ge 0} \sum_{i,j} \mathbf{T}_{ij} \,
c(H^{\mathbf{z}}_i, \tilde{H}^{\mathbf{z}'}_j), \\
\text{s.t. } \sum_{j=1}^{m} \mathbf{T}_{ij} = m_i^{\mathbf{z}}, \quad \forall i \in \{1,\ldots,k\}.
\end{aligned}
\end{equation}

This relaxation yields a lower bound of the exact OT distance and admits a closed-form solution:
each representation $H^{\mathbf{z}}_i$ transports all its probability mass to the most similar
$\tilde{H}^{\mathbf{z}'}_j$ under the cosine distance.
The resulting transport plan naturally supports unequal-length alignments, making it well-suited for sequences with heterogeneous tokenizations.

\subsection{Role in XBridge}

The proposed OT alignment provides a principled mechanism for aligning heterogeneous representations without assuming positional correspondence.
Moreover, since the multilingual encoder is frozen during training, the relaxed OT objective anchors the alignment to encoder-defined semantic geometry, encouraging decoder-side representations to remain compatible with the multilingual encoder-decoder space.
Despite its simplicity, this approximation is sufficient for our setting, as the goal is semantic compatibility regularization rather than exact distribution matching.

\section{Details for Training Data}
\label{details_training}

\paragraph{Translation Data in Stage~1}
We sample English-centric translation pairs from OPUS-100~\cite{zhang-etal-2020-improving} and filter the off-target pairs, with 50k samples per translation direction. For XBridge, we further translate English sentences into other languages $L_y$ using NLLB-200-3.3B to construct trilingual \textit{x-en-y} data. To mitigate translation noise in generation, we train XBridge using \textit{y-en-x}, where the encoder processes translated sentences and the decoder processes natural sentences.

\paragraph{Multilingual Reasoning Data and Multilingual Abstractive Summarization Data in Stage~2 and Stage~3}
We extract multilingual reasoning data from \citet{ruan-etal-2025-layalign}, which contains 30K samples per language across ten languages (the same as in Section~\ref{language_text}). 
We extract multilingual abstractive summarization data from XL-Sum~\cite{hasan-etal-2021-xl}. XL-Sum contains imbalanced multilingual data, and we have set the data upper limit to 30K.
For XBridge, we additionally construct bilingual responses using NLLB-200-3.3B.

\begin{figure*}[t]
\centering
  \includegraphics[width=0.9\linewidth]{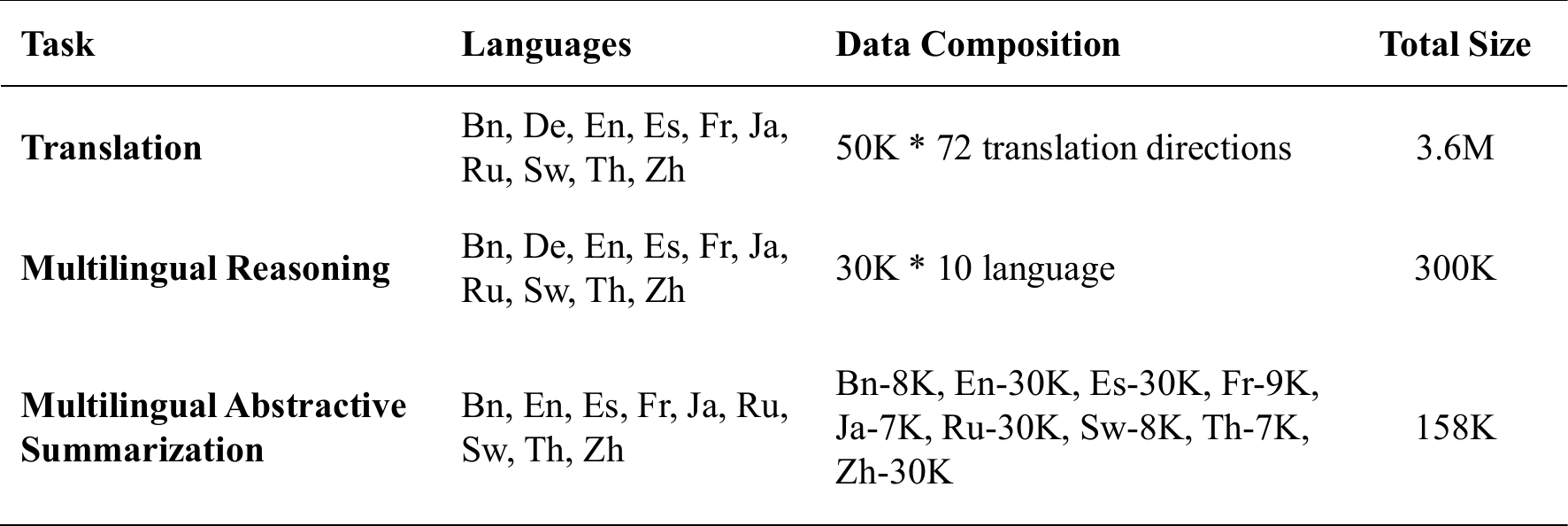}
  \caption{Statistics of training datasets used in different stages.}
  \label{statistics}
\end{figure*}

\paragraph{Data Statistics}
Figure~\ref{statistics} presents detailed data statistics for the training data.




\section{Detailed Results}
\label{detailed_results}

Table~\ref{table_detailed_flores}, Table~\ref{table_detailed_mgsm} and Table~\ref{table_detailed_xlsum} present detailed BLEU scores on FLORES-101 (COMET scores in Table~\ref{table_detailed_flores_comet}), accuracy on MGSM, and multilingual Rouge-L on XL-Sum for the main experiments in Section~\ref{main_experiment_text}. 

Table~\ref{table_detailed_ablation_flores}, Table~\ref{table_detailed_ablation_mgsm} and Table~\ref{table_detailed_ablation_xlsum} present results for the ablation study in Section~\ref{ablation_text}. 

Table~\ref{language_statistics} presents the included untuned languages and corresponding language codes. Table~\ref{table_detailed_untuned} presents detailed results for untuned language generalization in Section~\ref{untuned_language_text}. 

Table~\ref{cross_lingual_generation} presents BLEU scores for cross-lingual generation on FLORES-101, and Table~\ref{language_on_demand_generation} presents Accuracy for language-on-demand generation on MGSM in Section~\ref{cross_lingual_text}.

Table~\ref{table_detailed_m2m100_flores} and Table~\ref{table_detailed_m2m100_mgsm} present results for LLaMA3-8B composed with M2M100-1.2B in Section~\ref{m2m100_text}. 

Table~\ref{table_detailed_different_size_nmt_flores} and Table~\ref{table_detailed_different_size_nmt_mgsm} present results for MetaMath-7B composed with NLLB-200 in different sizes (600M vs. 1.3B) in Section~\ref{nmt_size_text}.

\section{Supplementary Analysis}
\label{addtional_analysis}

\subsection{Efficiency Analysis}
\label{efficiency_text}

We compare the training and inference efficiency of XBridge with SFT (LLM-only), MindMerger (Encoder-LLM), and the cascaded Translate-Test pipeline in Table~\ref{table_efficiency}. XBridge introduces only a limited training overhead despite the additional encoder and decoder, due to its parameter-efficient design.
For inference, XBridge is slower than the LLM-only method due to the additional decoding for multilingual generation, but it remains faster than the cascaded Translate-Test pipeline.
Overall, XBridge trades moderate computational cost for improved multilingual generation quality and robustness, while avoiding the inefficiency and error accumulation of cascaded systems.

\begin{table}[t]
\centering
\small
\begin{tabular}{l|cc}
\toprule
\textbf{System} & \textbf{Training} & \textbf{Inference} \\ \midrule
SFT                     & 1.0x              & 1.0x               \\
Translate-Test          & -                 & 0.55x              \\
MindMerger              & 1.42x             & 0.85x              \\
XBridge                 & 0.91x             & 0.66x              \\ \bottomrule
\end{tabular}
\caption{Relative speed comparison.}
\label{table_efficiency}
\end{table}

\subsection{Case Study on MGSM}
\label{case_study_text}

\begin{figure*}[htbp]
  \includegraphics[width=\linewidth]{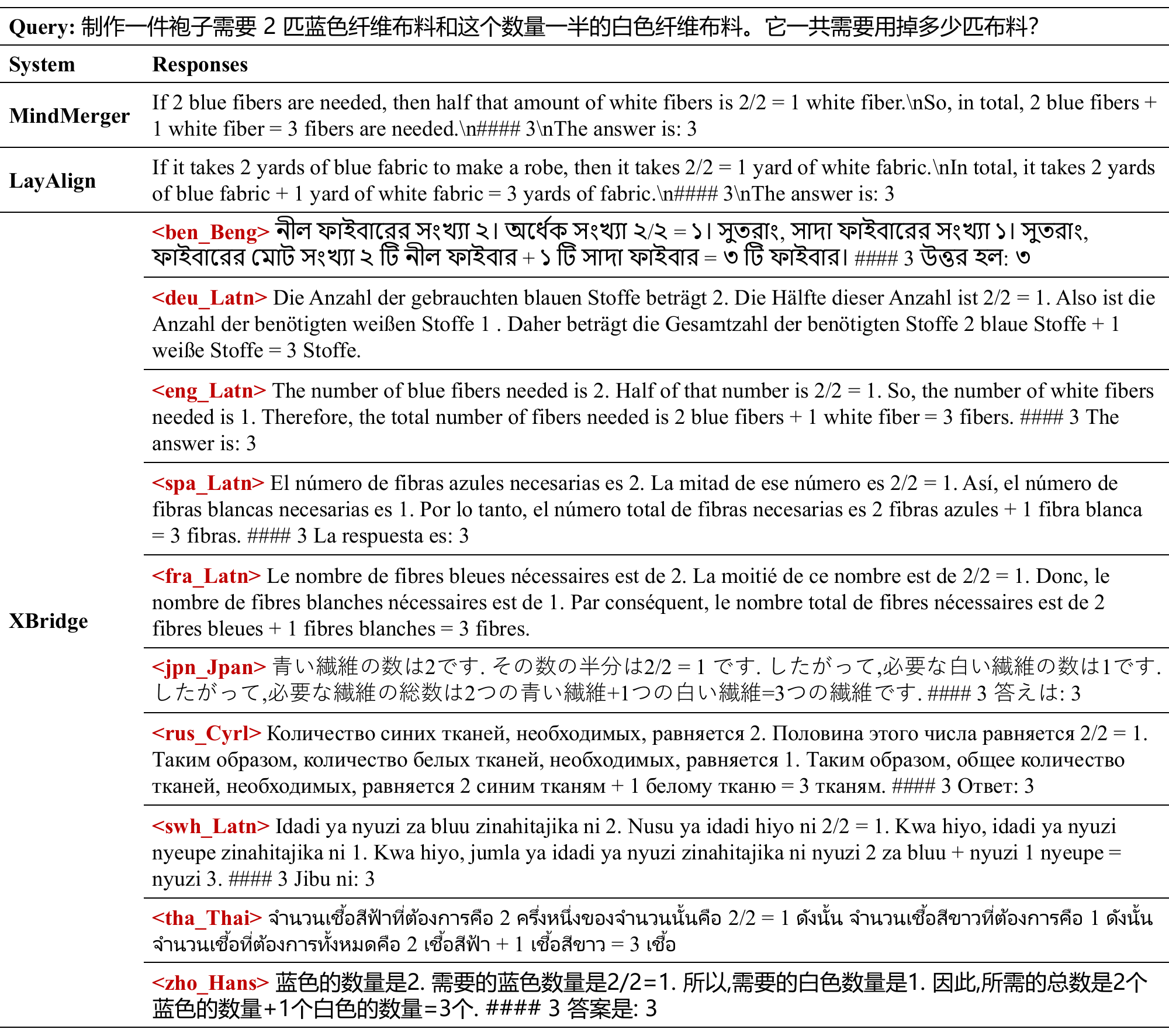}
  \caption{Case study on multilingual reasoning. Red text indicates the language token fed to the decoder, which controls the target generation language.}
  \label{case_study}
\end{figure*}

In the case study, we compare the outputs of MindMerger, LayAlign, and XBridge in Figure~\ref{case_study}.
MindMerger and LayAlign adopt encoder-augmented architectures, which only enable multilingual-to-English processing. As a result, the generated responses are always in English, which is less friendly for multilingual users who expect outputs in their native languages.
In contrast, XBridge supports controllable target-language generation by explicitly specifying the decoder language token, allowing the model to produce responses in different languages as required. This demonstrates the advantage of the compositional encoder-LLM-decoder design in providing flexible multilingual generation.

\subsection{Multilingual Commonsense Reasoning}
\label{csqa_text}

\begin{figure*}[t]
\centering
\includegraphics[width=0.8\linewidth]{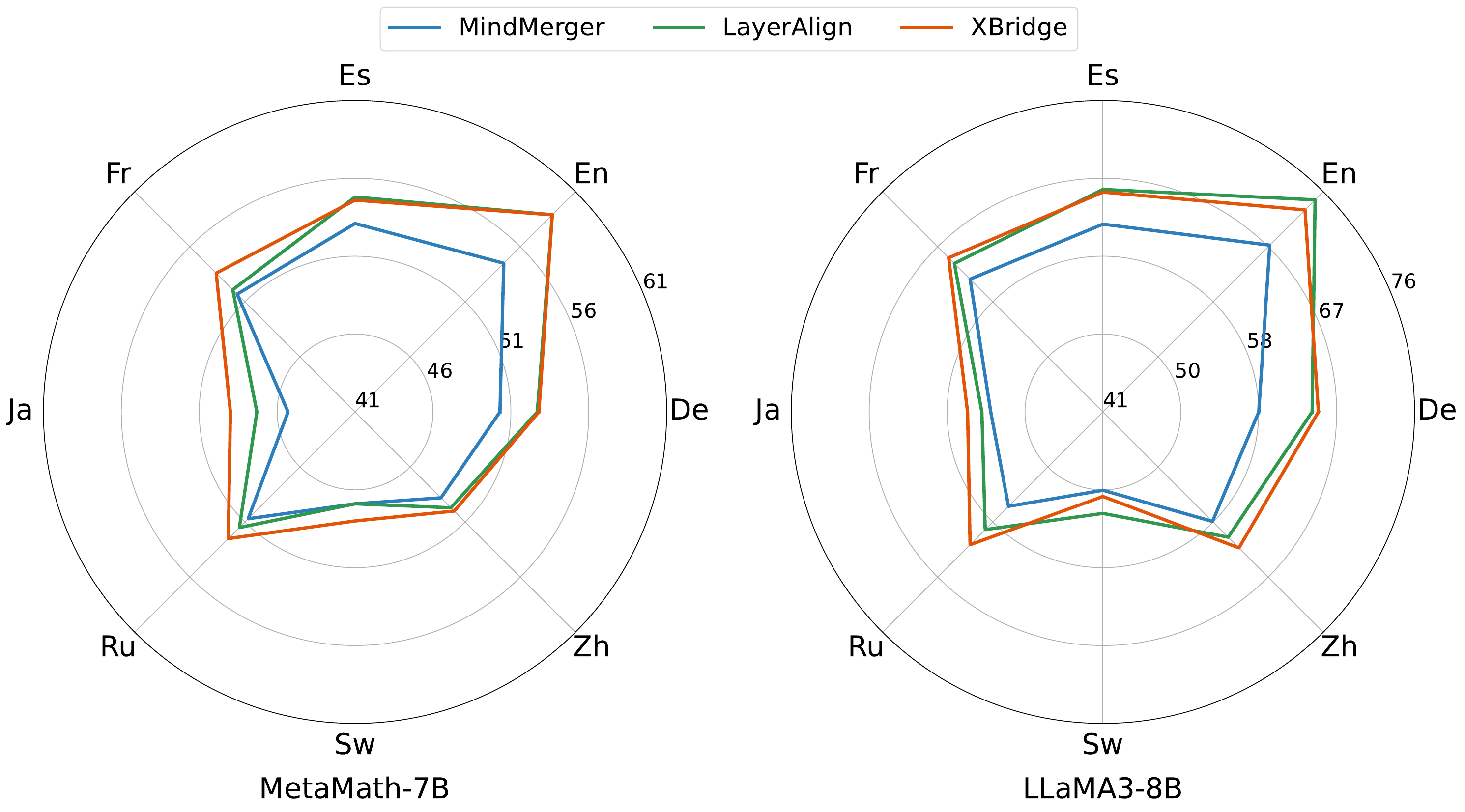}
\caption{Radar plot comparison on X-CSQA.}
\label{csqa}
\end{figure*}

We further extend XBridge to X-CSQA~\cite{lin-etal-2021-common} to evaluate both cross-lingual understanding and logical reasoning.
We report results on MetaMath-7B and LLaMA3-8B across eight languages in Figure~\ref{csqa}. XBridge outperforms MindMerger and LayAlign on MetaMath-7B across most languages, yielding the highest average accuracy.
Notably, the gains are more pronounced on non-English languages, while English performance is preserved.
This suggests that XBridge effectively enhances multilingual understanding without degrading the base model’s reasoning ability, and achieves more balanced multilingual reasoning performance across languages.

\begin{table*}[t]
\centering
\small
\begin{tabular}{cclcclcc}
\cmidrule(lr){1-2} \cmidrule(lr){4-5} \cmidrule(lr){7-8}
\textbf{ISO 639-1} & \textbf{Language} &  & \textbf{ISO 639-1} & \textbf{Language} &  & \textbf{ISO 639-1} & \textbf{Language}   \\ \cmidrule(lr){1-2} \cmidrule(lr){4-5} \cmidrule(lr){7-8}
Af                 & Akrikaans         &  & It                 & Italian           &  & Pl                 & Polish            \\
Ar                 & Arabic            &  & Ka                 & Georgian          &  & Ps                 & Pashto            \\
Az                 & Azerbaijani       &  & Kk                 & Kazakh            &  & Pt                 & Portuguese        \\
Cs                 & Czech             &  & Km                 & Khmer             &  & Ro                 & Romanian          \\
El                 & Modern Greek      &  & Ko                 & Korean            &  & Sl                 & Slovenian         \\
Et                 & Estonian          &  & Lt                 & Lithuanian        &  & Sv                 & Swedish           \\
Fa                 & Persian           &  & Lv                 & Latvian           &  & Ta                 & Tamil             \\
\end{tabular}
\caption{The included 41 untuned languages and corresponding language codes.}
\label{language_statistics}
\end{table*}

\begin{table*}[t]
\centering
\small
\begin{tabular}{lcccccccccc}
\toprule
\textbf{System}              & Bn-En & En-Bn & De-En & En-De & Es-En & En-Es & Fr-En & En-Fr & Ja-En & En-Ja \\ \midrule
\textbf{NLLB-200-1.3B}       & 37.78          & 32.83          & 46.23          & 39.91          & 35.37          & 31.35          & 47.14          & 51.33          & 29.60          & 19.07          \\ \midrule
\multicolumn{11}{c}{\textit{\textbf{MetaMath-7B}}}                                                                                                                                                     \\ \midrule
\textbf{MetaMath-7B}         & 1.46           & 0.67           & 34.36          & 19.42          & \textbf{36.33} & 27.42          & 15.54          & 11.64          & \textbf{27.62} & 16.76          \\
\textbf{MindMerger}          & 30.76          & -              & 40.05          & -              & 28.77          & -              & 38.65          & -              & 22.50          & -              \\
\textbf{LayerAlign}          & 30.91          & -              & 39.43          & -              & 30.16          & -              & 40.48          & -              & 22.36          & -              \\
\textbf{XBridge (Ours)}      & \textbf{35.47} & \textbf{29.23} & \textbf{41.42} & \textbf{35.39} & 30.51          & \textbf{28.88} & \textbf{42.14} & \textbf{49.49} & 24.52          & \textbf{19.60} \\ \midrule
\multicolumn{11}{c}{\textit{\textbf{LLaMA3-8B}}}                                                                                                                                                       \\
\bottomrule

\end{tabular}
\caption{Detailed BLEU scores on FLORES-101 for stage~1. "X" denotes all languages except for English. We bold the best scores for each LLM group.
}
\label{table_detailed_flores}
\end{table*}

\begin{table*}[t]
\centering
\small
\begin{tabular}{lcccccccccc} \toprule
\textbf{System}              & Bn-En & En-Bn & De-En & En-De & Es-En & En-Es & Fr-En & En-Fr & Ja-En & En-Ja \\ \midrule
\textbf{NLLB-200-1.3B}       & 88.30          & 85.67          & 88.80          & 86.12          & 86.72          & 85.68          & 88.83          & 87.05          & 87.05          & 86.45          \\ \midrule
\multicolumn{11}{c}{\textit{\textbf{MetaMath-7B}}}                                                                                                                                                     \\ \midrule
\textbf{MetaMath-7B}         & 50.85 & 35.10 & 86.25 & 76.59 & 84.39 & 78.49 & 86.84 & 78.45 & 82.62 & 76.35          \\
\textbf{MindMerger}          & 86.87          & -              & 87.81          & -              & 85.52          & -              & 87.64          & -              & 84.94          & -              \\
\textbf{LayerAlign}          & 87.10          & -              & 87.77          & -              & 85.75          & -              & 87.77          & -              & 84.78          & -              \\
\textbf{XBridge (Ours)}      & \textbf{88.03} & \textbf{84.98} & \textbf{88.17} & \textbf{85.31} & \textbf{86.16} & \textbf{85.26} & \textbf{88.07} & \textbf{86.53} & \textbf{86.04} & \textbf{85.99} \\ \midrule
\multicolumn{11}{c}{\textit{\textbf{LLaMA3-8B}}}                                                                                                                                                       \\
\bottomrule
\end{tabular}
\caption{Detailed COMET scores on FLORES-101 for stage~1. "X" denotes all languages except for English. We bold the best scores for each LLM group.
}
\label{table_detailed_flores_comet}
\end{table*}
\begin{table*}[t]
\small
\centering
\begin{tabular}{lccccccccccc}
\toprule
\textbf{System}     & \textbf{Bn}    & \textbf{De}    & \textbf{En}    & \textbf{Es}    & \textbf{Fr}    & \textbf{Ja}    & \textbf{Ru}    & \textbf{Sw}    & \textbf{Th}    & \textbf{Zh}    & \textbf{Avg}   \\ \midrule
\multicolumn{12}{c}{\textit{\textbf{MetaMath-7B}}}                                                                                                                                                             \\ \midrule
\textbf{SFT}                 & 30.00          & 40.40          & 48.00          & 39.60          & 39.20          & 32.40          & 39.60          & 35.20          & 32.80          & 36.00          & 37.32          \\
\textbf{Translate-Test} & 34.80 & 49.20 & 53.20 & 47.60 & 40.80 & 45.20 & 49.20 & 43.20 & 32.40 & 46.00 & 44.16 \\
\textbf{MindMerger}          & 43.60          & \textbf{57.20} & \textbf{67.20} & 59.20          & \textbf{60.00} & 45.60          & \textbf{57.60} & 49.20          & 42.40          & 48.80          & 53.08          \\
\textbf{LayerAlign}          & 40.40          & 54.80          & 62.80          & 59.20          & 52.40          & \textbf{51.60} & 56.40          & \textbf{53.20} & \textbf{47.20} & 51.20          & 52.92          \\
\textbf{XBridge\_LLM (Ours)} & \textbf{48.40} & 54.80          & \textbf{67.20} & \textbf{60.80} & 57.20          & 47.20          & \textbf{57.60} & 46.40          & 45.60          & \textbf{52.80} & \textbf{53.80} \\
\textbf{XBridge\_Dec (Ours)} & 47.20          & 52.80          & 66.80          & 60.00          & 51.60          & 44.80          & 56.00          & 45.60          & 44.40          & 52.00          & 52.12          \\
\bottomrule
\end{tabular}
\caption{Detailed accuracy results on MGSM. "XBridge-LLM" denotes English reasoning outputs from the LLM, while "XBridge-Dec" denotes multilingual reasoning outputs via the external decoder. We bold the best scores for each LLM group.}
\label{table_detailed_mgsm}
\end{table*}
\begin{table*}[t]
\small
\centering
\begin{tabular}{lcccccccccc}    \toprule
\textbf{System}         & \textbf{Bn}    & \textbf{En}    & \textbf{Es}    & \textbf{Fr}    & \textbf{Ja}    & \textbf{Ru}    & \textbf{Sw}    & \textbf{Th}    & \textbf{Zh}    & \textbf{Avg}   \\ \midrule
\multicolumn{11}{c}{\textit{\textbf{MetaMath-7B}}}                                                                                                                                                \\ \midrule
\textbf{SFT}            & 16.79          & \textbf{33.27} & \textbf{23.68} & \textbf{28.42} & 29.20          & \textbf{25.42} & 27.01          & 18.68          & 20.16          & 24.74          \\
\textbf{MindMerger}     & 17.93          & 27.27          & 21.03          & 26.05          & 28.11          & 22.25          & 26.49          & 19.20          & 18.56          & 22.99          \\
\textbf{LayerAlign}     & 17.94 & 28.84 & 21.49 & 26.31 & 27.55 & 22.22 & 26.29 & 20.09 & 19.02 & 23.30  \\
\textbf{XBridge (Ours)} & \textbf{18.93} & 31.66          & 22.75          & 27.96          & \textbf{32.12} & 24.64          & \textbf{28.90} & \textbf{20.70} & \textbf{24.12} & \textbf{25.75} \\ \midrule
\multicolumn{11}{c}{\textit{\textbf{LLaMA3-8B}}}                                                                                                                                                  \\ \midrule
\textbf{SFT}            & \textbf{20.87} & 33.59          & 23.84          & \textbf{30.18} & 30.42          & 26.12          & 28.73          & 22.51          & 21.72          & 26.44          \\
\bottomrule
\end{tabular}
\caption{Detailed multilingual Rouge-L results on XL-Sum. For XL-Sum, since the baselines produce English summaries only, we translate them into target languages using NLLB-200-1.3B for comparison. We bold the best scores for each LLM group.}
\label{table_detailed_xlsum}
\end{table*}
\begin{table*}[t]
\centering
\small
\begin{tabular}{lcccccc}
\toprule
\textbf{Variants}    & \textbf{Lrl-En} & \textbf{En-Lrl} & \textbf{Hrl-En} & \textbf{En-Hrl} & \textbf{X-En}  & \textbf{En-X}  \\ \midrule
\textbf{MetaMath-7B} & 2.87            & 1.05            & 26.50           & 17.36           & 18.62          & 11.92          \\
\textbf{w/o OT}      & 35.33           & 22.20           & 32.98           & 25.30           & 33.76          & 24.27          \\
\textbf{w/o Decoder} & \textbf{36.22}  & -               & \textbf{35.01}  & -               & \textbf{35.41} & -              \\ \midrule
\textbf{XBridge}     & 33.93           & \textbf{25.56}  & 31.98           & \textbf{29.37}  & 32.63          & \textbf{28.10} \\ \bottomrule
\end{tabular}
\caption{Ablation results on FLORES-101 for stage~1. "Lrl" and "Hrl" denote low- and high-resource languages, respectively. Following~\citet{shilanguage}, we treat Bn, Sw, and Th as low-resource languages, and the remaining as high-resource ones. "X" denotes all languages except for English. We bold the best scores for each LLM group.
}
\label{table_detailed_ablation_flores}
\end{table*}

\begin{table*}[t]
\centering
\small
\begin{tabular}{lcccccc}
\toprule
\textbf{Variants}         & \textbf{Lrl-LLM} & \textbf{Lrl-Dec} & \textbf{Hrl-LLM} & \textbf{Hrl-Dec} & \textbf{Avg-LLM} & \textbf{Avg-Dec} \\ \midrule
\textbf{MetaMath-7B}      & 5.60              & -                 & 50.06             & -                 & 36.72             & -                 \\ 
\textbf{SFT}              & 32.67             & -                 & 39.31             & -                 & 37.32             & -                 \\ \midrule
\textbf{w/o OT}           & 48.27             & 47.20             & 56.23             & 54.57             & 53.84             & 52.36             \\
\textbf{w/o Decoder}      & 48.27             & -                 & 56.91             & -                 & 54.32             & -                 \\
\textbf{w/o Stage 1}      & 29.33             & 28.13             & 52.51             & 51.37             & 45.56             & 44.40             \\
\textbf{Joint Stage 2\&3} & 43.07             & 42.53             & 48.63             & 47.26             & 46.96             & 45.84             \\ \midrule
\textbf{XBridge}          & \textbf{48.67}    & \textbf{47.87}    & \textbf{57.49}    & \textbf{55.09}    & \textbf{54.84}    & \textbf{52.92}    \\ \bottomrule
\end{tabular}
\caption{Ablation accuracy results on MGSM. "Lrl" and "Hrl" denote low- and high-resource languages, respectively. "-LLM" denotes English reasoning outputs from the LLM, while "-Dec" denotes multilingual reasoning outputs via the external decoder. We bold the best scores.
}
\label{table_detailed_ablation_mgsm}
\end{table*}

\begin{table*}[t]
\centering
\small
\begin{tabular}{lcccccccccccc}
\toprule
\textbf{Variants} & \textbf{Bn}    & \textbf{En}    & \textbf{Es}    & \textbf{Fr}    & \textbf{Ja}    & \textbf{Ru}    & \textbf{Sw}    & \textbf{Th}    & \textbf{Zh}    & \textbf{Lrl}   & \textbf{Hrl}   & \textbf{Avg}   \\ \midrule
\textbf{SFT}      & 16.79          & \textbf{33.27} & \textbf{23.68} & \textbf{28.42}          & 29.20          & \textbf{25.42} & 27.01          & 18.68          & 20.16          & 20.83          & 26.69          & 24.74          \\
\textbf{w/o OT}   & 18.55          & 31.49          & 22.68          & 27.86          & 32.05          & 24.27          & 28.35          & 20.27          & 23.81          & 22.39          & 27.03          & 25.48          \\ \midrule
\textbf{XBridge}  & \textbf{18.85} & 31.50 & 22.84 & 28.18 & \textbf{32.26} & 24.36          & \textbf{28.57} & \textbf{21.08} & \textbf{24.40} & \textbf{22.83} & \textbf{27.26} & \textbf{25.78} \\ \bottomrule
\end{tabular}
\caption{Ablation Rouge-L results of XL-Sum on MetaMath-7B. "Lrl" and "Hrl" denote low- and high-resource languages, respectively. We bold the best scores.}
\label{table_detailed_ablation_xlsum}
\end{table*}

\begin{table*}[t]
\centering \small
\begin{tabular}{c|cccc|cccc} \toprule
\multirow{2}{*}{\textbf{Languages}} & \multicolumn{4}{c|}{\textbf{X$\rightarrow$En}}                                                         & \multicolumn{4}{c}{\textbf{En$\rightarrow$X}}                                                         \\
             & \textbf{NLLB} & \textbf{MetaMath} & \textbf{w/o OT} & \textbf{XBridge} & \textbf{NLLB} & \textbf{MetaMath} & \textbf{w/o OT} & \textbf{XBridge} \\ \midrule
\textbf{Af}  & 58.33                  & 17.64                & 53.69           & \textbf{54.90}          & 42.69                  & 12.79                & 37.30           & \textbf{41.75}          \\
\textbf{Ar}  & 42.42                  & 24.59                & \textbf{40.61}  & 38.20                   & 33.13                  & 10.94                & 16.81           & \textbf{26.37}          \\
\textbf{Az}  & 28.25                  & \textbf{28.97}       & 25.35           & 23.57                   & 19.39                  & \textbf{16.23}       & 12.25           & 16.18                   \\
\textbf{Cs}  & 41.17                  & 1.34                 & \textbf{40.47}  & 38.54                   & 35.85                  & 0.32                 & 23.39           & \textbf{30.03}          \\
\textbf{El}  & 37.68                  & 3.59                 & \textbf{37.19}  & 35.43                   & 28.16                  & 0.80                 & 19.07           & \textbf{24.34}          \\
\textbf{Et}  & 38.22                  & 29.84                & \textbf{33.99}  & 33.30                   & 29.77                  & 9.62                 & 19.61           & \textbf{26.28}          \\
\bottomrule
\end{tabular}
\caption{Detailed results on 41 untuned languages, composing MetaMath-7B with NLLB-200-1.3B. We evaluate cross-model mapping quality on FLORES-101, following the main experiments. "X" denotes all languages except for English. We bold the best scores for the LLM group.}
\label{table_detailed_untuned}
\end{table*}
\begin{table*}[t]
\small
\centering
\begin{tabular}{c|cccccccccc|c} \toprule
    & Bn    & De    & En    & Es    & Fr    & Ja    & Ru    & Sw    & Th    & Zh    & Avg   \\ \midrule
Bn  & -     & 22.47 & 37.09 & 20.56 & 31.36 & 16.69 & 21.03 & 22.85 & 13.62 & 17.42 & 22.57 \\
De  & 24.74 & -     & 45.75 & 25.57 & 39.82 & 18.73 & 26.51 & 26.34 & 14.82 & 19.81 & 26.90 \\
En  & 28.42 & 35.45 & -     & 29.59 & 49.38 & 20.12 & 30.57 & 34.68 & 17.09 & 23.11 & 29.82 \\
Es  & 22.21 & 23.42 & 32.00 & -     & 33.74 & 16.82 & 21.78 & 22.01 & 12.93 & 17.57 & 22.50 \\
Fr  & 24.26 & 28.62 & 46.10 & 26.64 & -     & 17.75 & 25.90 & 26.99 & 14.47 & 19.96 & 25.63 \\
Ja  & 20.76 & 20.05 & 27.63 & 18.59 & 26.94 & -     & 18.82 & 19.76 & 12.60 & 17.16 & 20.26 \\
Ru  & 22.89 & 25.65 & 37.08 & 23.41 & 34.80 & 17.85 & -     & 23.96 & 13.91 & 18.82 & 24.26 \\
Sw  & 23.16 & 25.36 & 44.73 & 22.37 & 36.35 & 17.44 & 23.27 & -     & 14.24 & 18.18 & 25.01 \\
Th  & 21.59 & 20.98 & 30.61 & 19.34 & 29.39 & 16.37 & 19.78 & 21.86 & -     & 17.12 & 21.89 \\
Zh  & 20.36 & 19.78 & 24.89 & 18.90 & 27.12 & 15.89 & 18.74 & 20.21 & 12.94 & -     & 19.87 \\ \midrule
Avg & 23.15 & 24.64 & 36.21 & 22.77 & 34.32 & 17.52 & 22.93 & 24.30 & 14.07 & 18.79 & 23.87 \\ \bottomrule
\end{tabular}
\caption{BLEU scores for cross-lingual generation of XBridge, composing LLaMA3-8B with NLLB-200-1.3B. Rows denote the source language and columns denote the target language. The source text is encoded by the multilingual encoder, the LLM produces an English translation, and the decoder generates the target-language text conditioned on the target-language token. For \textit{X$\rightarrow$En} directions, we directly evaluate the LLM outputs.}
\label{cross_lingual_generation}
\end{table*}

\begin{table*}[t]
\centering \small
\begin{tabular}{c|cccccccccc|c} \toprule
\textbf{MGSM} & \textbf{Bn} & \textbf{De} & \textbf{En} & \textbf{Es} & \textbf{Fr} & \textbf{Ja} & \textbf{Ru} & \textbf{Sw} & \textbf{Th} & \textbf{Zh} & \textbf{Avg} \\ \midrule
\textbf{Bn}   & 46.00       & 44.40       & 45.20       & 45.60       & 40.80       & 42.80       & 44.00       & 46.00       & 44.00       & 44.80       & 44.36        \\
\textbf{De}   & 58.40       & 56.40       & 58.40       & 59.20       & 53.20       & 54.80       & 58.40       & 58.00       & 58.40       & 59.20       & 57.44        \\
\textbf{En}   & 63.60       & 62.80       & 63.20       & 63.60       & 59.60       & 61.60       & 62.80       & 63.20       & 61.20       & 64.00       & 62.56        \\
\textbf{Es}   & 60.00       & 57.60       & 61.60       & 61.20       & 55.20       & 56.80       & 60.40       & 60.80       & 57.60       & 60.80       & 59.20        \\
\textbf{Fr}   & 58.80       & 58.40       & 60.40       & 60.80       & 57.20       & 56.80       & 60.80       & 59.60       & 58.00       & 60.00       & 59.08        \\
\textbf{Ja}   & 49.20       & 47.20       & 49.20       & 50.00       & 43.60       & 46.80       & 48.80       & 48.80       & 48.00       & 48.80       & 48.04        \\
\textbf{Ru}   & 57.60       & 58.40       & 59.60       & 60.00       & 55.20       & 57.60       & 58.40       & 58.40       & 56.00       & 58.00       & 57.92        \\
\bottomrule     
\end{tabular}
\caption{Accuracy for langauge-on-demand generation of XBridge, composing LLaMA3-8B with NLLB-200-1.3B. Rows denote the query language and columns denote the response language. The query is first processed by the encoder, the LLM produces an English response, and the decoder then generates the target language response via the target language token.}
\label{language_on_demand_generation}
\end{table*}

\begin{table*}[t]
\small  \centering
\begin{tabular}{lcccccc}
\toprule
\textbf{System}         & \textbf{Lrl-En} & \textbf{En-Lrl} & \textbf{Hrl-En} & \textbf{En-Hrl} & \textbf{X-En}  & \textbf{En-X}  \\ \midrule
\textbf{M2M100-1.2B}    & 28.00           & 22.25           & 33.11           & 32.55           & 31.41          & 29.12          \\ \midrule
\textbf{LLaMA3-8B}      & 32.03           & 17.43           & 36.77           & 32.32           & 35.19          & 27.36          \\
\textbf{MindMerger}     & 31.37           & -               & 32.67           & -               & 32.24          & -              \\
\textbf{LayerAlign}     & 31.28           & -               & 32.47           & -               & 32.08          & -              \\
\textbf{XBridge (Ours)} & \textbf{32.24}  & \textbf{21.66}  & \textbf{35.86}  & \textbf{28.67}  & \textbf{34.65} & \textbf{26.33} \\ \bottomrule
\end{tabular}
\caption{Detailed FLORES-101 translation results for LLaMA3-8B composed with M2M100-1.2B. "Lrl" and "Hrl" denote low- and high-resource languages, respectively. "X" denotes all languages except for English. We bold the best scores for the LLM group.}
\label{table_detailed_m2m100_flores}
\end{table*}

\begin{table*}[t]
\centering
\small
\begin{tabular}{lccccccccccc} \toprule
\textbf{System}       & \textbf{Bn}    & \textbf{De}    & \textbf{En}    & \textbf{Es}    & \textbf{Fr}    & \textbf{Ja}    & \textbf{Ru}    & \textbf{Sw}    & \textbf{Th}    & \textbf{Zh}    & \textbf{Avg}   \\ \midrule
\textbf{SFT}                   & 38.80          & 46.40          & 54.00          & 54.80          & 44.40          & 38.80          & 45.20          & 40.80          & 44.00          & 49.20          & 45.64          \\
\textbf{Translate-Test} & 40.40 & 56.00 & 62.00 & 54.80 & 54.80 & 51.60 & 57.20 & 46.80 & 35.20 & 48.00 & 50.68 \\
\textbf{MindMerger}            & \textbf{52.40} & 55.60          & 62.40          & 55.60          & \textbf{59.60} & 50.40          & 58.40          & 47.60          & 50.80          & 56.40          & 54.92          \\
\textbf{LayerAlign}            & 51.20          & \textbf{60.00} & 65.20          & 58.00          & 55.60          & 50.00          & \textbf{59.20} & \textbf{49.20} & 50.80          & 54.00          & 55.32          \\
\textbf{XBridge\_LLM   (Ours)} & 49.20          & 57.20          & \textbf{66.00} & \textbf{60.80} & 52.80          & \textbf{54.40} & 54.80          & 47.20          & \textbf{52.80} & \textbf{60.00} & \textbf{55.52} \\
\textbf{XBridge\_Dec (Ours)}   & 48.40          & 53.20          & 62.80          & 59.20          & 48.40          & 53.20          & 54.00          & 44.40          & 48.40          & 58.80          & 53.08           \\ \bottomrule
\end{tabular}
\caption{Detailed MGSM results for LLaMA3-8B composed with M2M100-1.2B. "XBridge-LLM" denotes English reasoning outputs from the LLM, while "XBridge-Dec" denotes multilingual reasoning outputs via the external decoder. We bold the best scores for the LLM group.}
\label{table_detailed_m2m100_mgsm}
\end{table*}
\begin{table*}[t]
\small
\centering
\begin{tabular}{lcccccc}
\toprule
\textbf{System}         & \textbf{Lrl-En} & \textbf{En-Lrl} & \textbf{Hrl-En} & \textbf{En-Hrl} & \textbf{X-En}  & \textbf{En-X}  \\ \midrule
\textbf{NLLB-200-600M}  & 33.59           & 26.43           & 34.83           & 29.52           & 34.42          & 28.49          \\
\textbf{NLLB-200-1.3B}  & 37.09           & 28.86           & 37.72           & 32.08           & 37.51          & 31.00          \\ \midrule
\textbf{MetaMath-7B}    & 2.87            & 1.05            & 26.50           & 17.36           & 18.62          & 11.92          \\
\textbf{XBridge (600M)} & 31.20           & 23.42           & 29.46           & 26.55           & 30.04          & 25.50          \\
\textbf{XBridge (1.3B)} & \textbf{35.26}  & \textbf{26.91}  & \textbf{32.42}  & \textbf{31.24}  & \textbf{33.37} & \textbf{29.80} \\ \bottomrule
\end{tabular}
\caption{Detailed FLORES-101 translation results for MetaMath-7B composed with NLLB-200 in different sizes (600M vs. 1.3B). "Lrl" and "Hrl" denote low- and high-resource languages, respectively. "X" denotes all languages except for English. We bold the best scores for the LLM group.}
\label{table_detailed_different_size_nmt_mgsm}
\end{table*}

\begin{table*}[t]
\centering \small
\begin{tabular}{lccccccccccc} \toprule
\textbf{System}              & \textbf{Bn}    & \textbf{De}    & \textbf{En}    & \textbf{Es}    & \textbf{Fr}    & \textbf{Ja}    & \textbf{Ru}    & \textbf{Sw}    & \textbf{Th}    & \textbf{Zh}    & \textbf{Avg}   \\ \midrule
\textbf{SFT}                 & 38.80          & 46.40          & 54.00          & 54.80          & 44.40          & 38.80          & 45.20          & 40.80          & 44.00          & 49.20          & 45.64          \\
\textbf{XBridge\_LLM (600M)} & 47.60          & \textbf{56.00} & 65.60          & 56.00          & 56.00          & \textbf{47.20} & 56.00          & \textbf{50.80} & \textbf{49.20} & 50.00          & 53.44          \\
\textbf{XBridge\_Dec (600M)} & 45.20          & 52.80          & 65.60          & 55.60          & 51.60          & 44.40          & 54.40          & \textbf{50.80} & 46.00          & 49.20          & 51.56          \\
\textbf{XBridge\_LLM (1.3B)} & \textbf{48.40} & 54.80          & \textbf{67.20} & \textbf{60.80} & \textbf{57.20} & \textbf{47.20} & \textbf{57.60} & 46.40          & 45.60          & \textbf{52.80} & \textbf{53.80} \\
\textbf{XBridge\_Dec (1.3B)} & 47.20          & 52.80          & 66.80          & 60.00          & 51.60          & 44.80          & 56.00          & 45.60          & 44.40          & 52.00          & 52.12         \\ \bottomrule
\end{tabular}
\caption{Detailed FLORES-101 translation results for MetaMath-7B composed with NLLB-200 in different sizes (600M vs. 1.3B). "X" denotes all languages except for English. We bold the best scores for the LLM group.}
\label{table_detailed_different_size_nmt_flores}
\end{table*}

\end{document}